\documentclass[journal]{IEEEtran}
\usepackage{booktabs}
\usepackage{amssymb}
\usepackage{amsfonts}
\usepackage{bbding}
\usepackage{svg}
\usepackage{subfigure}
\usepackage[numbers]{natbib}

\ifCLASSOPTIONcompsoc
 \usepackage[caption=false,font=normalsize,labelfont=sf,textfont=sf]{subfig}
\else
 \usepackage[caption=false,font=footnotesize]{subfig}
\fi
\usepackage{dblfloatfix}
\usepackage{url}
\begin{document}

\def\UrlFont{\em}

\title{Exchanging Dual Encoder-Decoder: A New Strategy for Change Detection with Semantic Guidance and Spatial Localization}

\author{Sijie~Zhao,
        Xueliang~Zhang,~\IEEEmembership{Member,~IEEE,}
        Pengfeng~Xiao,~\IEEEmembership{Senior~Member,~IEEE,}
        and~Guangjun~He
\thanks{Sijie Zhao, Xueliang Zhang, and Pengfeng Xiao are with the Jiangsu Provincial Key Laboratory of Geographic
Information Science and Technology, Key Laboratory for Land Satellite Remote Sensing Applications of Ministry of Natural Resources, School of Geography and Ocean Science, Nanjing University, Nanjing 210023, China (e-mail: zsj@smail.nju.edu.cn; zxl@nju.edu.cn; xiaopf@nju.edu.cn).

Guangjun He is with the State Key Laboratory of Space-Ground Integrated
Information Technology, Space Star Technology Co., Ltd., Beijing 100095,
China (e-mail: hgjun\_2006@163.com).

Corresponding Author: X. Zhang. This work was supported by the National Natural Science Foundation of China (Grant No. 42071297), and the Fundamental Research Funds for the Central Universities under Grant 020914380119.

}
}

\maketitle

\begin{abstract}
Change detection is a critical task in earth observation applications. Recently, deep learning-based methods have shown promising performance and are quickly adopted in change detection. However, the widely used multiple encoder and single decoder (MESD) as well as dual encoder-decoder (DED) architectures still struggle to effectively handle change detection well. The former has problems of bitemporal feature interference in the feature-level fusion, while the latter is inapplicable to intraclass change detection and multiview building change detection. To solve these problems, we propose a new strategy with an exchanging dual encoder-decoder structure for binary change detection with semantic guidance and spatial localization. The proposed strategy solves the problems of bitemporal feature inference in MESD by fusing bitemporal features in the decision level and the inapplicability in DED by determining changed areas using bitemporal semantic features. We build a binary change detection model based on this strategy, and then validate and compare it with 18 state-of-the-art change detection methods on six datasets in three scenarios, including intraclass change detection datasets (CDD, SYSU), single-view building change detection datasets (WHU, LEVIR-CD, LEVIR-CD+) and a multiview building change detection dataset (NJDS). The experimental results demonstrate that our model achieves superior performance with high efficiency and outperforms all benchmark methods with F1-scores of 97.77\%, 83.07\%, 94.86\%, 92.33\%, 91.39\%, 74.35\% on CDD, SYSU, WHU, LEVIR-CD, LEVIR-CD+, and NJDS datasets, respectively. The code of this work will be available at \url{https://github.com/NJU-LHRS/official-SGSLN}.

\end{abstract}


\IEEEpeerreviewmaketitle

\section{Introduction}

\IEEEPARstart{C}{hange} detection is the process of identifying differences in the state of an object or phenomenon by observing it at different times~\cite{change_detection_definition}. It is crucial in applications such as urban expansion investigations~\cite{urban_expansion}, land use planning~\cite{lulc}, and disaster damage assessments~\cite{disaster_damage}. Binary change detection is the process of identifying changed objects of interest given binary labels, which is basic but of great significance in change detection. Binary change detection can be classified into intraclass change detection (ICCD) and specific-class change detection, where the former detects all categories of changed objects, and the latter detects specific categories of changed objects. Since binary labels only provide the change information of changed objects rather than their semantic information, ICCD faces great challenges in detecting multiple categories of changed objects. Building change detection occupies an important position in specific-class change detection, which is important for urban planning and monitoring illegal construction~\cite{edge}. According to the imaging angles of multitemporal remote sensing images, building change detection can be classified into single-view building change detection (SVBCD) with similar imaging angles and multiview building change detection (MVBCD) with large differences in imaging angles, where the latter is common and faces great challenges in very high resolution remote sensing images. In SVBCD, the edge parts of changed buildings are difficult to accurately detect due to factors such as building shadows and the dense distribution of buildings. In MVBCD, because of the different imaging angles of multitemporal remote sensing images, the same building has large spatial differences in the bitemporal images, leading to the confusion of real changes and thus false positives.

In recent years, deep learning methods have been quickly adopted in remote sensing due to the advent of massive remote sensing data and the rapid development of deep learning~\cite{ai_emergence_1,ai_emergency_2}. A large number of deep learning-based change detection methods have been developed for binary change detection~\cite{dsifn,hsal,change_mask,fccdn}. There are two types of neural networks widely used in binary change detection: multiple encoder and single decoder (MESD) network~\cite{fcef,dsifn} and dual encoder-decoder (DED) network~\cite{fccdn,rasr}. 

\begin{figure*}[!ht]
	\centering
		\includegraphics[width=\linewidth]{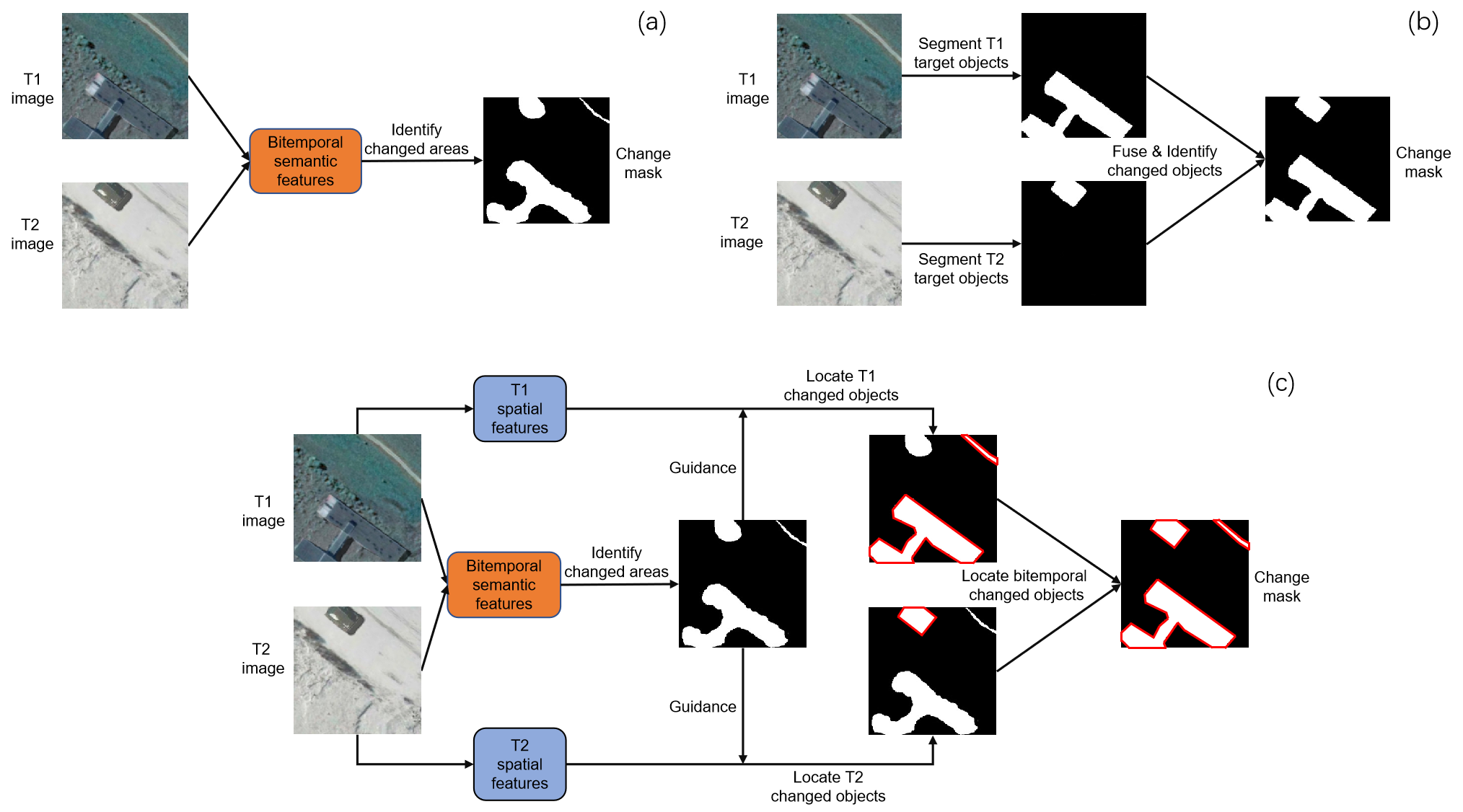}
	\caption{Illustration of the main idea of MESD, DED and EDED. (a) MESD: Changed areas can be roughly identified by bitemporal semantic features. (b) DED: The specific types of changed objects can be identified by comparing the segmentation results of bitemporal target objects. (c) EDED: Bitemporal changed objects can be identified by bitemporal semantic features and located by bitemporal spatial features, where the red edges denotes changed objects in each temporal phase.}
 \label{Fig:idea}
\end{figure*}

MESD consists of multiple encoders with shared weights and a single decoder for change detection. Bitemporal semantic features are extracted in multiple encoders and fused in the feature level in the single decoder to identify the changed areas, as shown in Figure \ref{Fig:idea} (a). This network suffers from problems in the feature-level fusion: when fusing bitemporal encoder features, the changed object features in one temporal phase could be contaminated by the background features in another phase at the same spatial location, leading to inferior performance of the network~\cite{fccdn}. As an example, Figure \ref{Fig:idea} (a) shows that the MESD can only identify the rough changed areas of changed objects.

DED consists of a dual encoder-decoder with shared weights and a single decoder for change detection. The input bitemporal remote sensing images are fed into the dual encoder-decoder to segment the target objects in each image. The bitemporal features are then fused in the decision level in the single decoder to detect changed objects, as shown in Figure \ref{Fig:idea} (b). DED solves the problems of bitemporal feature contamination in MESD by segmenting the target objects in each image and fusing them in the decision level to detect changed objects.

However, DED has two assumptions: the target objects in the bitemporal images can be segmented accurately, and the changes can be retrieved correctly by comparing the target objects~\cite{fccdn}. Therefore, DED faces great challenges in ICCD and MVBCD. In ICCD, there are multiple types of changes occurring in different classes of objects, while the binary labels only indicate the presence of changes without specifying the change types. Therefore, it is difficult for DED to segment the target objects without semantic labels of bitemporal images. As an example, Figure \ref{Fig:idea} (b) shows that DED detects the changed buildings while missing the changed roads. In MVBCD, due to the different imaging views, there are significant spatial discrepancies for the same object in bitemporal images. Since DED aims to segment the target objects in bitemporal images accurately, the spatial discrepancies of the same object will be mistaken as changed areas, leading to false positives.

Current neural networks also have other limitations for change detection: (1) Most change detection networks focus on the important parts within features in each temporal when fusing bitemporal features, neglecting the important parts across the bitemporal features; and (2) Most change detection networks have a large number of parameters and require huge computational resources, resulting in time-consuming training and inference.

To address the aforementioned challenges, we propose a new strategy with exchanging dual encoder-decoder (EDED) structure for binary change detection, as shown in Figure \ref{Fig:idea} (c). 
EDED have the same structure with DED except for a channel exchange module, which leads to a new strategy for change detection. In EDED, spatial features in each temporal phase are extracted in the shallow layers of the dual encoder and half-exchanged, which makes features in each temporal branch both contain bitemporal features. Therefore, changed areas can be determined as guidance using bitemporal semantic features in the deep layers of the dual encoder. Next, based on the changed areas, the T1 changed objects are located accurately using T1 spatial features. The changed objects in phase T2 are located in the same way. Finally, all changed objects can be located accurately when fusing bitemporal decoder features in the decision level. As an example, Figure \ref{Fig:idea} (c) shows that EDED can successfully detect the changed buildings and roads in the T1 image and the changed buildings in the T2 image.

EDED solves the problem of bitemporal feature concatenation in MESD by separately locating the changed objects in each image and fusing them in the decision level. Moreover, EDED can overcome the limitations in DED by determining the changed areas to identify all types of changed objects in the ICCD and distinguish the false change caused by view differences by using bitemporal semantic features in the MVBCD.

We also design a temporal fusion attention module (TFAM) and a half-convolution unit (HCU), in which the former focuses on the important parts across bitemporal features using temporal information, and the latter reduces the parameters and computation of conventional convolution to 1/4.

Based on these works, we propose a semantic guidance and spatial localization network (SGSLN) for binary change detection. The main contributions of this study are as follows: 

\begin{enumerate}

    \item We propose an exchanging dual encoder-decoder backbone as a new strategy for binary change detection. It uses bitemporal semantic features to determine changed areas and spatial features in each temporal phase to locate changed objects in the corresponding phase, thus locating all changed objects by fusing bitemporal changed object features. In this way, it addresses the issues of bitemporal feature contamination in MESD and overcomes the inapplicability in ICCD and MVBCD scenarios in DED.

    \item We design a temporal fusion attention module to fuse bitemporal features effectively. By comparing bitemporal features in the temporal dimension, the model can identify the important part across bitemporal features and achieve effective feature fusion.

    \item We design a half-convolution unit to reduce the parameters and computation of the network. It has 1/4 the number of parameters and computation of conventional convolution and can facilitate feature reuse, allowing the model to achieve fast and robust training and inference.
\end{enumerate}

\section{Related Works}

\subsection{Classical Change Detection Methods}

Classical change detection algorithms mainly include layer arithmetic, post-classification change, direct classification, transformation, change vector analysis (CVA), and hybrid change detection~\cite{tcd}. 

Layer arithmetic method compare image radiance or derivative features numerically to identify changes. For example, Coulter et al.~\cite{la1} utilized regionally normalized NDVI measures to detect changes in vegetative land cover. While this approach is straightforward to apply, it often offers limited insight into the detected changes.

Post-classification change method is the process of overlaying thematic maps from different time periods to pinpoint changes. One of the most established and extensively employed change detection techniques is directly comparing land cover maps derived from satellite data~\cite{pc1,pc2}. This method offers a comprehensive thematic approach that can address specific queries about changes, making it applicable across various domains. Nevertheless, any error present in the input maps could be directly reflected in the resultant change map.

Direct classification method utilizes a multi-temporal data stack as input, classifying it using supervised or unsupervised techniques to establish a set of consistent land cover classes and detect changes in land cover transitions. For example, Chehata et al.~\cite{dc1} implemented a forest change detection system by employing unsupervised classification on multi-temporal imagery. This method only needs one classification stage and can provide an effective framework to mine a complicated time series. However, constructing training datasets for such a classification can be highly demanding, and unsupervised methods might not effectively capture subtle changes in magnitude~\cite{dc2}.

Data transformations, such as Principal Component Analysis (PCA) and Multivariate Alteration Detection (MAD), can be employed on a multi-temporal stack of remotely sensed images to emphasize variance between images and facilitate change identification. For example,  Doxani et al.~\cite{t1} found that implementing the MAD transformation on image-objects effectively highlighted changed objects in Very High-Resolution (VHR) imagery. Similarly, Chen et al.~\cite{t2} utilized the MAD transformation on image-objects to accentuate change. These transformations offer a useful approach for assessing changes in complex time series of images. However, their primary function is often to highlight changes, thus they should ideally be integrated into a hybrid change detection workflow. Notably, due to scene-specific features, locating changes within multiple components may prove challenging, particularly if the change is not distinctly represented~\cite{tcd}.

CVA is a technique for interpreting change by considering both its magnitude and direction. For example, Bruzzone and
Prieto~\cite{cva1} computed the change magnitude across all six Landsat spectral bands to evaluate the apparent extent of change. Analyzing the magnitude and direction of change vectors can provide insights into the types of the changes. However, this approach can also introduce ambiguity because the change vector itself can be repositioned within the feature space while preserving the same magnitude and direction measurements~\cite{cva2}. Consequently, there is a possibility that various thematic changes might yield identical measures of magnitude and direction.

Hybrid change detection method employs multiple comparison methods simultaneously to enhance the comprehension of detected changes. At a fundamental level, it can be conceptualized as a two-stage process: change localization and change identification. For example, Doxani et al.~\cite{t1} tackled urban change detection in VHR imagery by first utilized a MAD transform to highlight changed areas, and then applied a knowledge-based classification to filter and classify the results. This methodology reflects a research trend that incorporates multiple stages of change comparison to address specific challenges~\cite{tcd}.

\begin{figure*}[!ht]
	\centering
		\includegraphics[width=0.8\linewidth]{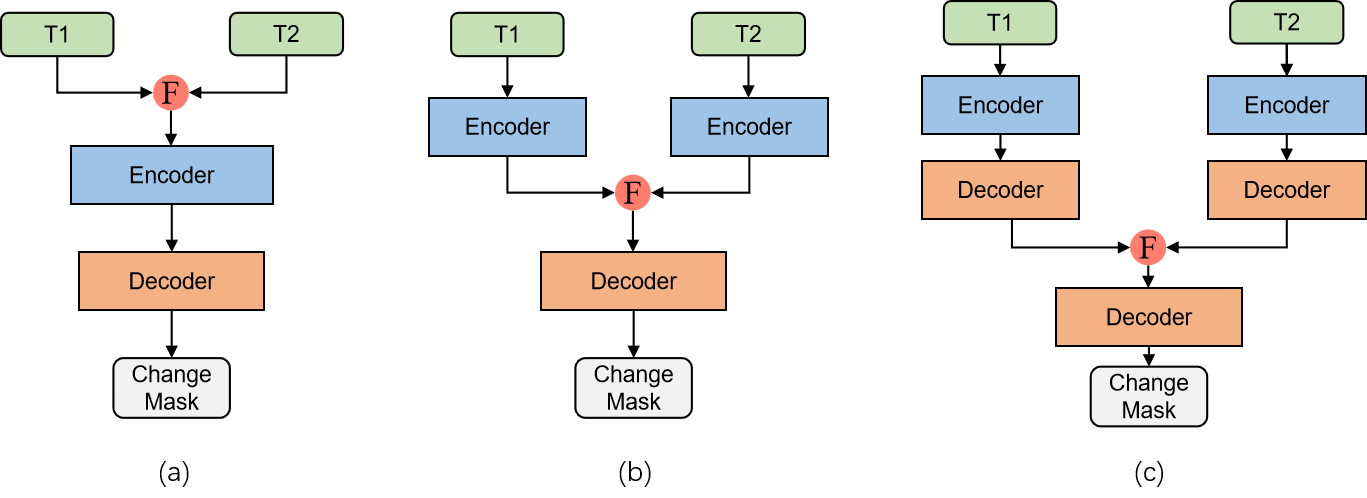}
	\caption{Illustration of the architecture of SED, MESD and DED. T1 and T2 denote bitemporal images, F denotes a fusion module. (a) SED: Single encoder-decoder architecture. (b) MESD: Multiple encoders and single decoder architecture. (c) DED: Dual encoder-decoder architecture.}
 \label{Fig:related_backbone}
\end{figure*}

\subsection{Deep Learning Based Change Detection Method}
Deep learning offers the capability to extract useful features and make accurate decisions by leveraging extensive sets of remote sensing images, which allows deep learning-based methods to outperform traditional methods in many remote sensing applications. There are mainly three widely used architectures in deep learning-based change detection models: single encoder-decoder, multiple encoder and single decoder networks, and dual encoder-decoder.

\subsubsection{Single Encoder-Decoder}
Single encoder-decoder (SED) uses a single encoder-decoder architecture to generate a change map from concatenated or difference images of bitemporal images, as shown in Figure \ref{Fig:related_backbone} (a). Papadomanolaki et al.~\cite{duc} proposed a deep learning framework for urban change detection, which combines U-Net~\cite{unet} for feature extraction and LSTMs~\cite{lstm} for temporal modeling.  Peng et al.~\cite{orsi} used UNet++\_MSOF~\cite{etecd} as the backbone, in which the spatial and channel attention strategies are used in the upsampling unit and the difference features are used to refine the results. Zheng et al.~\cite{clnet} proposed a convolutional neural network in which the U-Net~\cite{unet} structure is used as the backbone and cross-layer blocks are embedded to incorporate multiscale features and multilevel context information. Most SED networks are modified from networks for single-image semantic segmentation. Since bitemporal images are fused as one input before being fed into the networks, early layers fail to provide informative deep features of individual raw images, which consequently results in change maps with broken object boundaries and poor object internal compactness~\cite{dsifn}.

\subsubsection{Multiple Encoder and Single Decoder}
Multiple encoder and single decoder (MESD) feeds bitemporal images and their difference image if necessary into multiple encoders to extract features, which are then merged and upsampled in a single decoder to generate the change map, as shown in Figure \ref{Fig:related_backbone} (b). MESD can extract the information of images in each temporal by Siamese network structure while maintaining a similar number of parameters and computation as SED by weight-sharing. Hou et al.~\cite{hrtn} proposed a change detection method that uses triple encoders and multiscale modules to extract features, which are then fused with upsampling and distance computation to produce the change map.  Zhu et al.~\cite{lulc} proposed a change detection network using a global hierarchical sampling mechanism to address the imbalanced training sample problem with insufficient samples. As bitemporal features may contaminate each other in the feature-level fusion, how to fuse them effectively becomes a challenge. Zhang
et al.~\cite{dsifn} concatenated and fused bitemporal features with channel and spatial attention strategies. Zhang et al.~\cite{olcd} fused bitemporal features with a convolution enhancement approach and self-attention in spatial and channel dimensions. Chen
et al.~\cite{edge} fused bitemporal features with a feature differential enhancement module, in which both local and global information are exploited and beneficial for bitemporal feature fusion. However, these fusion methods focus on the enhancement of the features themselves, ignoring the temporal information between bitemporal features. At the same time, bitemporal features interfere with each other at the positions of changed objects in the feature-level fusion, making it difficult to detect the changed objects accurately.

\subsubsection{Dual Encoder-Decoder}
Dual encoder-decoder (DED) feeds bitemporal images into a dual encoder-decoder to segment target objects in each image, which are then fused in a single change decoder to generate a change map, as shown in Figure \ref{Fig:related_backbone} (c). DED networks are commonly used in multitask change detection and semantic change detection, where both segmentation labels and change labels are needed for training the model. However, DED networks are rarely applied in binary change detection since DED depends on the supervision of two segmentation branches. Chen et al.~\cite{fccdn} demonstrated that the DED structure with only binary change labels supervised outperforms MESD in change detection, and further improved the performance of DED by utilizing a self-supervised learning (SSL) strategy. SSL enables supervision of a dual segmentation branch by making bitemporal segmentation results the pseudo labels for each other with a specific loss function, in which the unchanged part should be similar, and the changed part should be different between bitemporal segmentation maps.  Liang et al.~\cite{rasr} adopted the same network structure and SSL strategy as~\cite{fccdn}, with additional deep supervision modules to train the network better and relation-aware modules to enhance features.

However, DED depends on the accurate segmentation of bitemporal images and obtains change maps by comparing segmentation maps~\cite{fccdn}, making it challenging to adapt to ICCD and MVBCD. ICCD with multiple change types and binary labels makes it difficult for DED to segment target objects of bitemporal images. MVBCD with different imaging angles between bitemporal images makes DED mistake the differences caused by different imaging angles of bitemporal target objects as changed areas.

\section{Methodology}

\subsection{Overall Structure of the Proposed Network}
\label{section:3.1}

\begin{figure*}[!ht]
	\centering
		\includegraphics[width=0.9\linewidth]{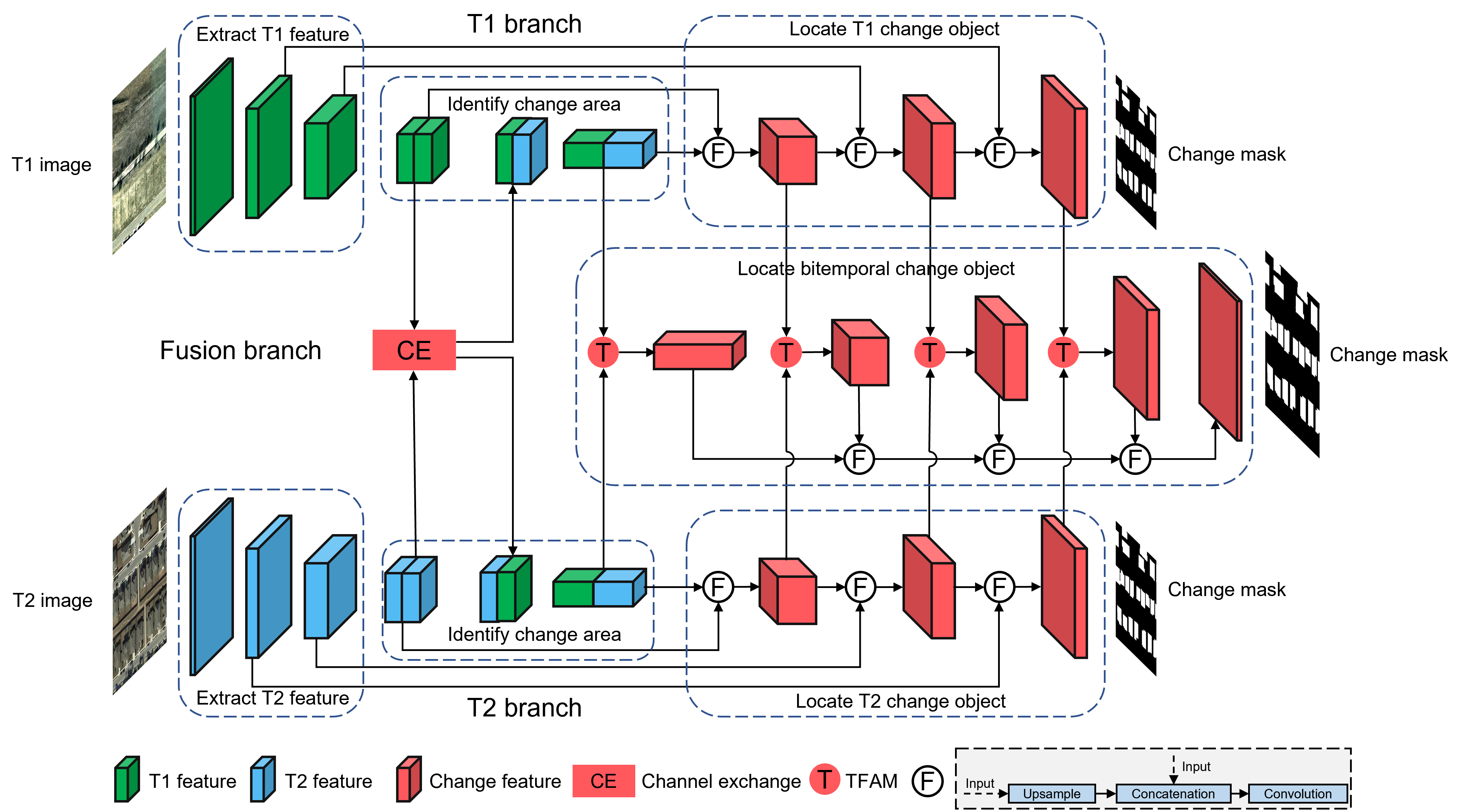}
	\caption{Overall structure of the SGSLN. CE denotes the channel exchange module, and TFAM denotes the temporal fusion attention module.}
 \label{Fig:SGSLN}
\end{figure*}

SGSLN consists of an exchanging dual encoder-decoder backbone (EDED) with two weight-sharing encoders, two weight-sharing decoders, and a fusion decoder, as shown in Figure \ref{Fig:SGSLN}. Each encoder and decoder is composed of a series of half convolution units (HCUs) and convolutional block attention modules (CBAMs~\cite{cbam}) for effective feature extraction.

The dual weight-sharing encoder extracts spatial features of bitemporal images in the shallow layers. Bitemporal encoder features are then half-exchanged, which means features in each encoder both contain bitemporal features. After that, bitemporal encoder features are passed to deep layers of the dual encoder so that the changed areas can be roughly identified in each encoder by exploiting the bitemporal semantic features, which provide guidance for subsequent changed object localization.

Based on the changed areas, the decoder in the T1 branch can precisely locate T1 changed objects by using the spatial features in T1 encoder features when fusing T1 decoder features and T1 encoder features through skip connections. The decoder in T2 the branch can locate T2 changed objects precisely in the same way. The dual decoders in the bitemporal branches both generate a change mask with half the size of the input images, which are supervised with change labels to reduce the path length of gradient back-propagation and train the model effectively. As bitemporal changed objects are located in bitemporal decoder features, temporal fusion attention modules (TFAM) in the fusion branch are designed to determine the relatively important parts between bitemporal features, thus effectively fusing the bitemporal changed object features and locating all changed objects. The decoder in the fusion branch generates a change mask with the same size as the input images, which is the result of the SGSLN.

\subsection{Exchanging Dual Encoder-Decoder Backbone}
\label{section:3.2}

\begin{figure*}[!ht]
 \centering
 	\includegraphics[width=0.8\linewidth]{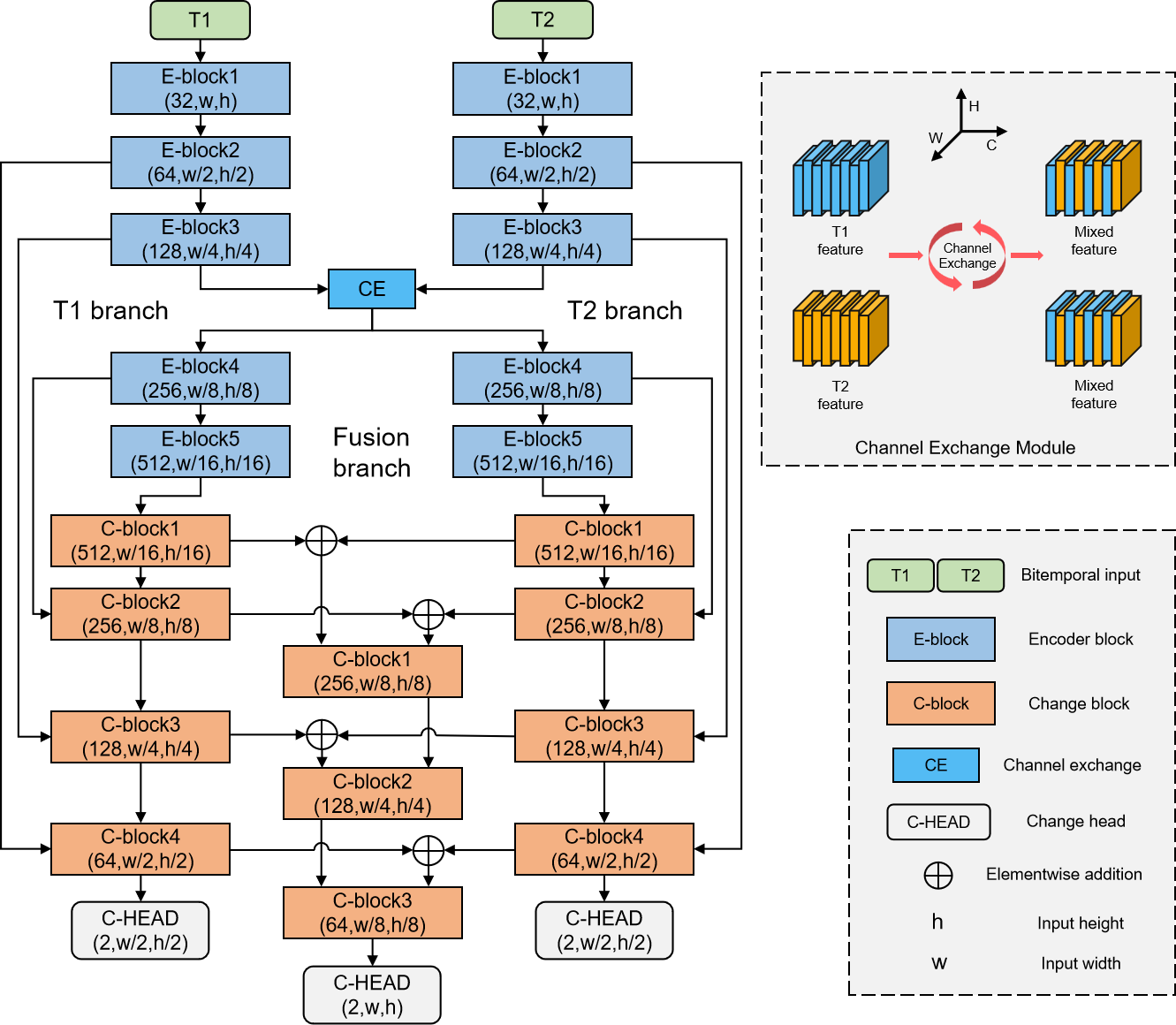}
 \caption{EDED backbone for change detection. T1 and T2 denote the bitemporal remote sensing images inputs, and CE denotes channel exchange module, which is shown in the top right corner.}
 \label{Fig:EDED}
\end{figure*}

\begin{figure}[!ht]
 \centering
 \subfigure[Encoder block 1]{
\begin{minipage}[b]{0.295\linewidth}
\centering
\includegraphics[width=0.4\linewidth]{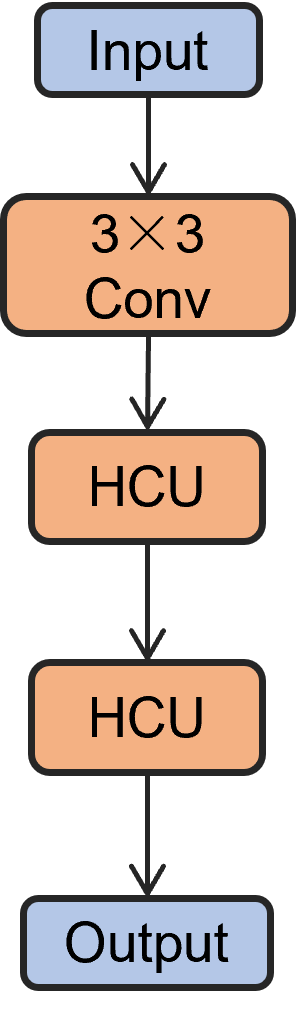}
\end{minipage}
}
\subfigure[Encoder block 2-5]{
\begin{minipage}[b]{0.295\linewidth}
\centering
\includegraphics[width=0.45\linewidth]{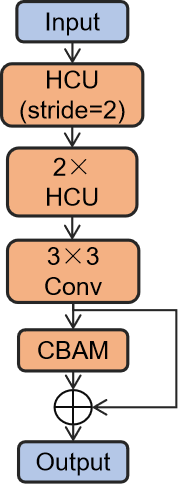}
\end{minipage}
}
\subfigure[Change block]{
\begin{minipage}[b]{0.295\linewidth}
\centering
\includegraphics[width=\linewidth]{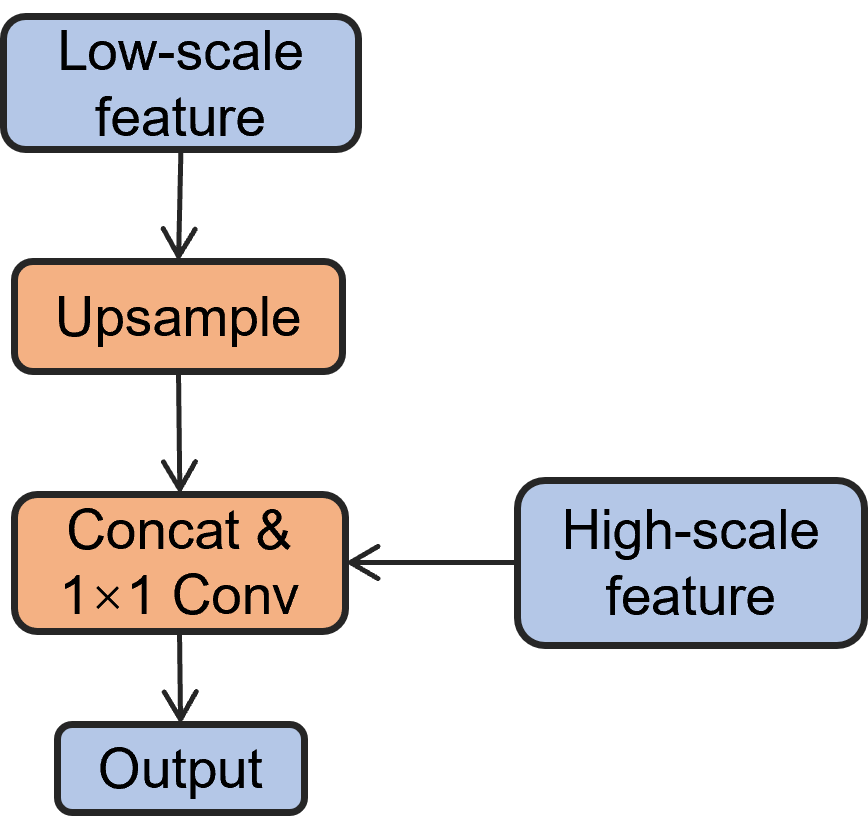}
\end{minipage}
}
 \caption{Details of encoder blocks and change blocks. HCU denotes the half convolution unit, 3×3 Conv refers to the convolution layer with kernel size=3, 1×1 Conv refers to the convolution layer with kernel size=1, and Concat refers to the concatenation of two features in the channel dimension.}
 \label{Fig:blocks}
\end{figure}

EDED and DED have the same structure except for a channel exchange module, which completely changes the strategy of the model for change detection, as shown in Figure \ref{Fig:EDED}. EDED and DED both have a dual encoder-decoder and a single decoder. The dual encoder-decoder in DED is used to segment the bitemporal changed objects separately, which means that the features in each encoder-decoder branch only contain single temporal information and ignore the connection between the bitemporal changed objects. In contrast, with the channel exchange module between the dual encoder-decoder in EDED, each branch contains bitemporal information after channel exchange, which means that bitemporal features are connected and each branch can determine the changed areas itself. Based on the changed areas, the features with only single temporal information before channel exchange contain rich spatial features, which can be used to refine changed areas and accurately locate the changed objects in the same temporal phase. Finally, all the changed objects can be precisely located by fusing bitemporal features. 

In EDED, bitemporal remote sensing images are fed into dual encoder blocks 1-3 to extract bitemporal features in each branch. The bitemporal features are then half-exchanged in the channel exchange module, which alternately exchanges half of the input bitemporal features in the channel dimension, as shown in the top right corner of Figure \ref{Fig:EDED}. The process of channel exchange module can be formulated as :
\begin{equation}
    T'_1, T'_2 = M * T_1 + (1-M) * T_2
\end{equation}
where $T_1$ and $T_2$ denote bitemporal features, $T'_1$ and $T'_2$ denote exchanged bitemporal features, and $M$ denotes the one-dimension exchange mask, in which the length is equal to the channel dimension size of bitemporal features and the values are filled with 0 and 1 alternately. In this way, each exchanged feature contains half of the bitemporal features, which means that each exchanged feature contains bitemporal semantic features of bitemporal remote sensing images. Therefore, dual encoder blocks 4-5 in the bitemporal branch can determine rough changed areas using the bitemporal semantic features.

Since the changed areas only contain part of the changed objects, they can only determine the approximate location of the changed objects but cannot completely detect the changed objects. Therefore, we use the spatial features of each temporal image to refine the changed areas. Taking the changed areas as guidance, the decoder in the T1 branch fuses the encoder features with the decoder features through skip connections, and uses the spatial features of the T1 changed objects in the encoder features to refine the changed areas, thus completely detecting the T1 changed objects. The decoder in the T2 branch completely detects the T2 changed objects in the same way. The dual decoders in bitemporal branches both generate change masks and are supervised by the change labels to train the dual branches better. Based on the bitemporal changed objects located in the bitemporal decoder, the decoder in the fusion branch can accurately locate all changed objects when fusing bitemporal decoder features and generate a change map as the result of the model.

The blocks in EDED are carefully designed to obtain a strong feature extraction ability while keeping lightweight. Encoder block 1 extracts bitemporal features of input bitemporal remote sensing images without downsampling, making the bitemporal features contain rich original information, as shown in Figure \ref{Fig:blocks} (a). Encoder blocks 2-5 first use a HCU with stride equals to 2 to downsample the input features, then use 3×3 convolution after two HCUs to ensure sufficient cross-channel interaction of features, and finally use CBAM to enhance the features, as shown in Figure \ref{Fig:blocks} (b). In this way, encoder blocks can achieve effective feature extraction with feature reuse of lightweight HCUs and feature enhancement of CBAMs, so that encoder blocks 1-3 have rich spatial features to locate bitemporal changed objects, and encoder blocks 4-5 have rich semantic features to determine the changed areas. Change blocks upsample the input low-resolution features and fuse them with high-resolution features, thus identifying multiscale changed objects using multiscale features, as shown in Figure \ref{Fig:blocks} (c).

EDED locates the changed objects separately and fuses them in the decision level, so that the changed object features in one temporal image will not be interfered with by the background feature in another temporal image at the same position when fusing bitemporal features, thus solving the bitemporal feature interference problems in MESD. Moreover, based on the semantic features of bitemporal images, EDED can determine changed areas of all categories of changed objects in ICCD, and distinguish real changes and pseudochanges caused by viewing angle differences in MVBCD, thus overcoming the inapplicability in ICCD and MVBCD scenarios in DED.

\begin{figure}[!ht]
 \centering
 \includegraphics[width=\linewidth]{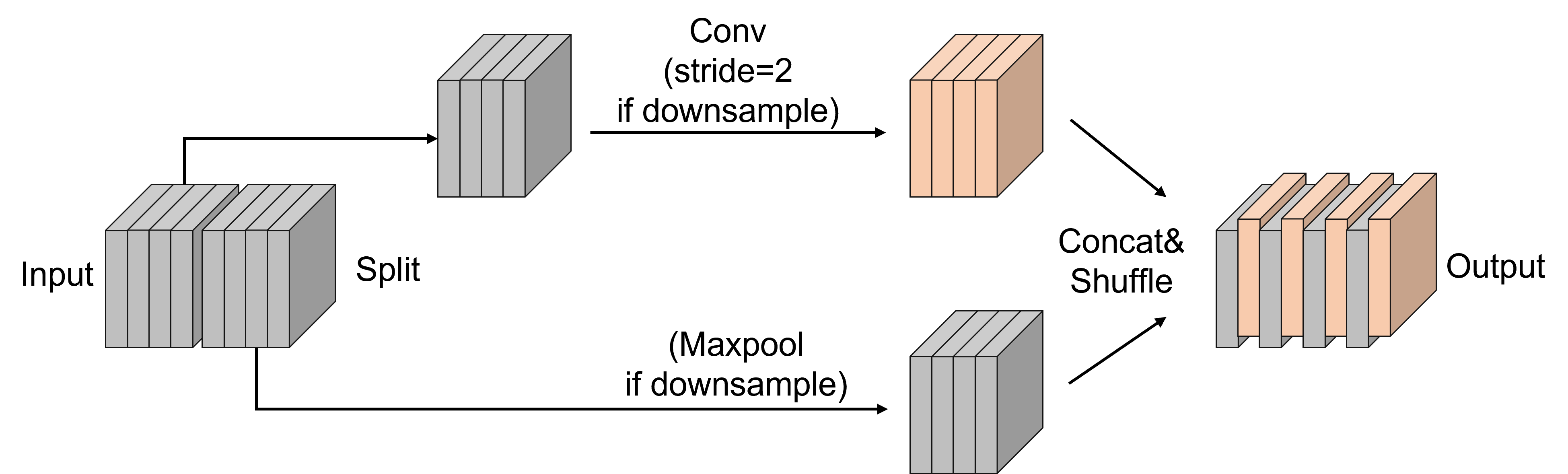}
 \caption{Illustration of the half-convolution unit. Half-features passed to convolution layers are concatenated and shuffled with another residual half-features.}
 \label{Fig:HCU}
\end{figure}

\subsection{Half Convolutional Unit}
\label{section:3.3}

we propose an HCU to replace the conventional convolution, which reduces the parameters and computation of the model and achieves effective feature extraction, as shown in Figure \ref{Fig:HCU}. The input features are split into two halves in the channel dimension, one of which is passed to convolution layers to be enhanced and the other serves as residual features. Features in two branches are then concatenated in the channel dimension and shuffled in an alternating arrangement, generating the output features. If the input features need to be downsampled, the stride of convolution is set to 2 and the residual features are downsampled using maxpool with stride equals to 2.

The shuffle operation can ensure sufficient cross-channel interaction, and retaining half of the input features is beneficial to gradient back-propagation and feature reuse~\cite{shuffle_v2}. In this way, HCU has only 1/4 of the parameters and computation of conventional convolution while maintaining a strong feature extraction ability, thus making the model lightweight and achieving effective feature extraction.

\subsection{Temporal Fusion Attention Module}
\label{section:3.4}

\begin{figure*}[!ht]
	\centering
		\includegraphics[width=\linewidth]{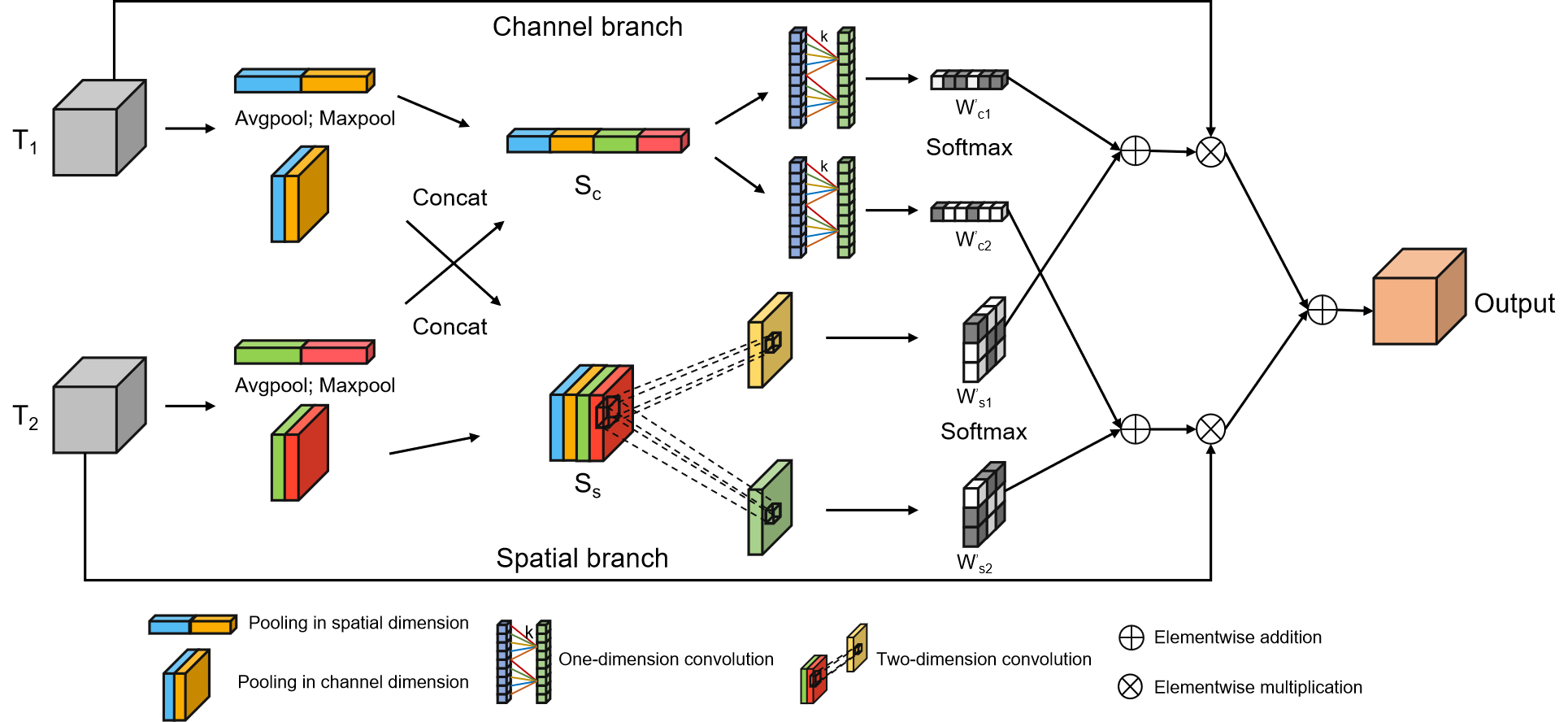}
	\caption{Illustration of the structure of TFAM. T1 and T2 refer to bitemporal features. Avgpool and Maxpool denote average pooling and max pooling in the channel and spatial dimensions, respectively. Concat denotes the concatenation of two features in the channel dimension.}
 \label{Fig:TFAM}
\end{figure*}

Bitemporal feature fusion methods in change detection models can be classified into simple fusion, convolution enhancement, and attention enhancement~\cite{fcef, snu,fccdn, dsifn, olcd}. Simple fusion method directly performs elementwise addition, subtraction, or concatenation on bitemporal features to fuse them~\cite{fcef, snu}. This method is susceptible to noise interference in bitemporal features, and it is difficult to achieve effective feature fusion. Convolution enhancement method enhances bitemporal features of multiple scales and semantic levels by applying various convolution operations, which reduces noise interference in bitemporal features. Then, it fuses bitemporal features using addition, subtraction, or concatenation~\cite{fccdn}. Attention enhancement method usually concatenates bitemporal features in the channel dimension and then achieves effective fusion using attention mechanisms~\cite{dsifn, olcd}. However, convolution enhancement method focuses on the enhancement of bitemporal features before fusion, and attention enhancement method focuses on the enhancement of bitemporal features after simple fusion. They both ignore the temporal information between bitemporal features.

To solve the above issues, we propose a TFAM to utilize temporal information for effective feature fusion, which is shown in Figure \ref{Fig:TFAM}. It uses channel and spatial attentions to determine the important parts in features and uses temporal information to determine the important parts between bitemporal features. In the channel branch, the input bitemporal features are passed through global pooling across the spatial dimension to aggregate spatial information. The aggregated process can be formulated as :
\begin{equation}
    S_c = Concat(Avg(T_1),Max(T_1),Avg(T_2),Max(T_2))
\end{equation}
where $S_c$ denotes the aggregated spatial features, $T_1$ and $T_2$ denote bitemporal features, $Avg(.)$ and $Max(.)$ denote global average pooling and global max pooling across spatial dimension, respectively. The aggregated spatial features are passed to two one-dimensional convolutions, which are the same as ECA modules~\cite{eca} to determine the bitemporal channel weights of the input bitemporal features. The two channel weights can be formulated as : 
\begin{equation}
    W_{c1}, W_{c2} = Conv_1(S_c), Conv_2(S_c)
\end{equation}
where $W_{c1}$ and $W_{c2}$ denote bitemporal channel weights, $Conv_1(.)$ and $Conv_2(.)$ denote one-dimension convolutions. Softmax is then used in bitemporal channel weights to make their summation equal to 1, which means comparing bitemporal weights to determine the higher value between them, thus determining the important parts between bitemporal features in the channel dimension. The softmax approach can be formulated as :
\begin{equation}
    W'_{c1}, W'_{c2} = \frac{e^{W_{c1}}}{e^{W_{c1}}+e^{W_{c2}}}, \frac{e^{W_{c2}}}{e^{W_{c1}}+e^{W_{c2}}}
\end{equation}
where $W'_{c1}$ and $ W'_{c2}$ denote output bitemporal channel weights. The bitemporal spatial weights $W'_{s1}$ and $ W'_{s2}$ are determined in the same way in the spatial branch, thus determining the important parts between bitemporal features in the spatial dimension. Bitemporal channel weights and bitemporal spatial weights are summarized to obtain bitemporal weights, which determine the important parts between bitemporal features. Finally, bitemporal weights are multiplied with bitemporal features and summarized to effectively fuse bitemporal features. The output can be formulated as :
\begin{equation}
    Output = (W'_{c1} + W'_{s1}) * T_1 + (W'_{c2} + W'_{s2}) * T_2
\end{equation}
where $Output$ denotes the fused features. As the summation of bitemporal weights is equal to 1, the useful parts between bitemporal features are retained while the useless parts are discarded, thus achieving effective feature fusion.

\section{Experimental Settings and Results}

We conducted experiments on three scenarios of six datasets to evaluate whether SGSLN is applicable for binary change detection: two for ICCD (CDD~\cite{cdd} and SYSU~\cite{sysu} datasets), three for SVBCD (WHU~\cite{whu}, LEVIR-CD~\cite{sta} and LEVIR-CD+ datasets), and one for MVBCD (NJDS~\cite{njds} dataset). We compared three versions of SGSLN with 18 state-of-the-art change detection methods. Three versions of SGSLN are SGSLN/128, SGSLN/256 and SGSLN/512. The number in the name indicates the maximum channel size of the encoder features in the model. The model with a larger channel number has a larger number of parameters and computation. 

\subsection{Datasets}

We offer a brief description of the experimental binary change detection datasets in Table \ref{datasets_introduction}. 

\subsubsection{Intraclass Change Detection}

The CDD dataset~\cite{cdd} consists of 11 pairs of season-varying Google Earth images covering various objects (such as buildings, roads, and vehicles) that change. The dataset excludes the changes caused by seasonal differences and brightness, which makes it challenging for the change detection algorithm. The dataset is cropped into patches of 256×256 pixels, with 10 000 patches for training, 3000 patches for validation, and 3000 patches for testing.

The SYSU dataset~\cite{sysu} contains images that capture various types of complex change scenes, such as road expansion, new urban buildings, vegetation change, suburban growth, and groundwork before construction. We split the data into training, validation, and test sets at a ratio of 6:2:2, following the same approach as~\cite{sysu}.

\subsubsection{Single-View Building Change Detection}

The WHU dataset~\cite{whu} consists of two-period aerial images acquired in 2012 and 2016, which contain various buildings with large-scale changes. Following the splitting approach used in~\cite{orsi}, we crop the dataset into nonoverlapping patches of 256×256 pixels and randomly split them into training/validation/test sets with a ratio of 7:1:2.

The LEVIR-CD dataset~\cite{sta} is a large-scale change detection dataset that contains very high-resolution (0.5 m/pixel) Google Earth images. The images capture various types of buildings that have changed for 5 to 14 years. The dataset focuses on building-related changes, such as building growth and decline. The bitemporal images are labeled by experts using binary masks (1 for change and 0 for unchanged). The dataset has a total of 31,333 individual change-building instances. Following~\cite{fccdn}, we crop the images into patches of 256×256 pixels with an overlap of 128 pixels on each side (horizontal and vertical) and split the samples into training/validation/test sets with a ratio of 7:1:2.

The LEVIR-CD+ dataset is an extension of the LEVIR-CD dataset. It contains 985 pairs of images acquired from 2002 to 2020, with approximately 80000 building instances. We follow the same splitting and cropping approach as the LEVIR-CD dataset on the LEVIR-CD+ dataset.

\subsubsection{Multiview Building Change Detection}

The NJDS dataset~\cite{njds} addresses the building height displacement issue in change detection. It contains bitemporal images of Nanjing City in 2014 and 2018, obtained from Google Earth. The images include different types of low-, middle-, and high-rise buildings. Following the same approach as~\cite{njds}, we crop the images into nonoverlapping patches of 256×256 pixels and randomly split them into training (540 pairs), validation (152 pairs), and testing sets (1,827 pairs).

\begin{table}[!ht]
\caption{Brief Introduction of The Experimental Datasets.}
\label{datasets_introduction}
\centering
\begin{tabular}{lcccc}
\toprule
Name & Resolution (m) & Image pairs & Image size (pixels) \\
\midrule
CDD & 0.03-1 & 16000 & 256×256 \\
SYSU & 0.5 & 20000 & 256 × 256 \\
WHU & 0.3 & 1 & 32207×15354 \\
LEVIR-CD & 0.5 & 637 & 1024×1024 \\
LEVIR-CD+ & 0.5 & 985 & 1024×1024 \\
NJDS & 0.3 & 1 & 14231×11381 \\
\bottomrule
\end{tabular}
\end{table}

\subsection{Benchmark Methods}
We compare the proposed method with change detection networks based on SED, MESD, and DED architectures to verify its effectiveness. The benchmark methods tested on the same dataset are based on the same splitting of dataset and use the same data, except that SFCCD~\cite{njds} additionally uses segmentation labels of bitemporal images.

In SED architecture-based networks, bitemporal remote sensing images are concatenated in the channel dimension and fed into a fully convolution-based network to obtain a change map. The compared change detection networks based on SED include U-Net~\cite{unet}, AttU-Net~\cite{aunet}, PSPNet~\cite{psp}, FC-EF~\cite{fcef}, UNet++\_MSOF~\cite{etecd}, and Intelligent-BCD~\cite{ibcd}.

In MESD architecture-based networks, bitemporal remote sensing images and their addition or subtraction are fed into multiple encoders to extract features and fused in a single decoder to obtain a change map. The compared change detection networks based on MESD include FC-Siam-Diff~\cite{fcef}, FC-Siam-Conc~\cite{fcef}, DTCDSTN~\cite{dtcdstn}, IFN~\cite{dsifn}, SNUNet~\cite{snu}, STANet~\cite{sta}, TransUNetCD~\cite{transunetcd}, and DARNet~\cite{darn}.

In DED architecture-based networks, bitemporal remote sensing images are fed into a dual encoder-decoder to segment bitemporal target objects. Bitemporal target object features are then fused in the change decoder to obtain a change map. The compared change detection networks based on DED include BiT~\cite{bit}, FCCDN~\cite{fccdn}, MTU-Net~\cite{mtu}, and SFCCD~\cite{njds}.

\subsection{Implementation Details}

\subsubsection{Data Augmentation}

We apply various data augmentation techniques in the training stage to enhance the generalization ability of the models. These techniques include random flipping (probability = 0.5), transposing (probability = 0.5), random shifting (probability = 0.3), random scaling (probability = 0.3), random rotation (probability = 0.3), and one of the following transformations with probability = 0.3: HSV shifting, Gaussian noise, brightness and contrast adjustment, gamma noise, embossing, and motion blur. We use Albumentations~\cite{albu} to implement all data augmentation methods with the default settings. Moreover, we randomly exchange the input order of the bitemporal images with probability = 0.5.

\subsubsection{Training and Inference}
\label{section:4.3.2}

We use PyTorch~\cite{pytorch} to implement the SGSLN and train it on 1 RTX A5000 GPU (24 GB memory). The batch size is 64 for our network. We adopt the binary cross entropy loss and dice coefficient loss as the loss function. AdamW~\cite{adam} is used as the optimizer with an initial learning rate of 0.001 and a weight decay of 0.001. For the learning rate adjustment scheduler, we reduce the learning rate by 0.1 if the F1-score of the validation set does not increase within 12 epochs. We train the network for 250 epochs and save the checkpoints with the highest F1-scores on the validation sets for testing. The choice of 250 epochs was made to ensure that the model receives sufficient training and has reached convergence. The first 30 epochs are skipped in the validation as the model is far from converging. On the LEVIR-CD, CDD, SYSU, and NJDS datasets, we initialize the models following PyTorch's default settings to keep the same parameter initial method with other change detection methods. As the pretrained model can improve robustness and accelerate the model to converge~\cite{upt}, following the parameter initialized method in~\cite{fccdn}, we use the pretrained model trained on the LEVIR-CD dataset to initialize the SGSLN in the experiments on the WHU and LEIVR-CD+ datasets.

\subsubsection{Evaluation Metrics}

We use precision (P), recall (R), F1-score and intersection over union (IoU) as the evaluation metrics for change detection. These metrics are widely used to measure the performance of change detection models. Precision measures the false positives in results while recall measures the false negatives. It is difficult
to achieve high precision and recall simultaneously. The F1 score is the harmonic mean of precision and recall, which can balance the trade-off by taking both metrics into account. The IoU is the ratio of the overlapping area between the predicted changed pixels and the changed pixels to the area of their union.

\subsection{Ablation Study}
\label{section:4.4.1}

To verify the effectiveness and superiority of EDED, we conduct comparative experiments among MESD, DED and EDED on three change detection datasets of different scenarios (SYSU dataset for ICCD, LEVIR-CD dataset for SVBCD and NJDS dataset for MVBCD). In addition, to verify the effectiveness of HCU and TFAM, we conduct ablation experiments on HCU and TFAM on LEVIR-CD dataset using EDED as the backbone.

EDED outperforms MESD and DED in all experimental change detection scenarios and achieves good performance, as shown in Table \ref{backbone_ablation}. The results indicate that in ICCD and SVBCD, DED performs better than MESD, but in the MVBCD scenario, DED performs worse than MESD. In the above three change detection scenarios, EDED performs better than MESD and DED, especially in ICCD and MVBCD. The results of MESD, DED and EDED further support the argument in Section \ref{section:3.2}.

\begin{figure*}[!ht]
\centering
\subfigure[]{
\begin{minipage}[b]{0.18\linewidth}
\includegraphics[width=1\linewidth]
{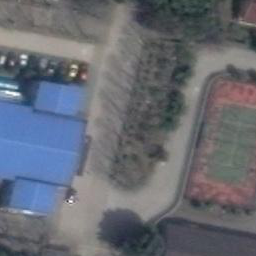} \\ \\
\includegraphics[width=1\linewidth]{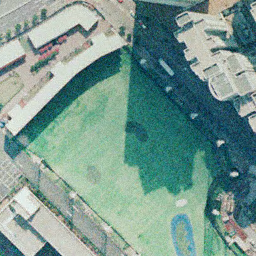}
\end{minipage}
}
\subfigure[]{
\begin{minipage}[b]{0.18\linewidth}
\includegraphics[width=1\linewidth]{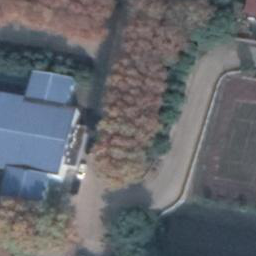} \\ \\
\includegraphics[width=1\linewidth]{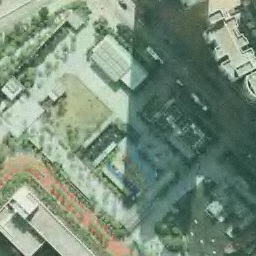}
\end{minipage}
}
\subfigure[]{
\begin{minipage}[b]{0.18\linewidth}
\includegraphics[width=1\linewidth]{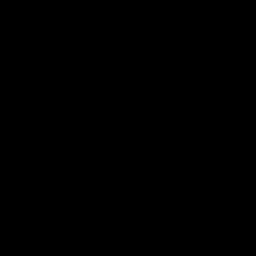} \\ \\
\includegraphics[width=1\linewidth]{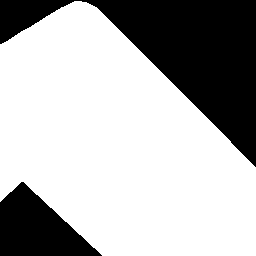}
\end{minipage}
}
\subfigure[]{
\begin{minipage}[b]{0.18\linewidth}
\includegraphics[width=1\linewidth]{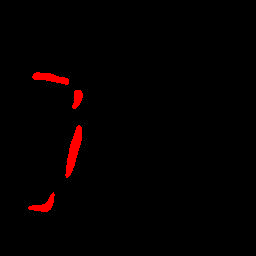} \\ \\
\includegraphics[width=1\linewidth]{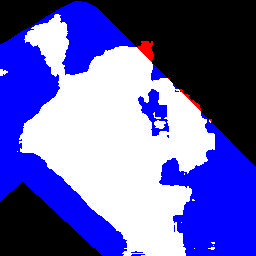}
\end{minipage}
}
\subfigure[]{
\begin{minipage}[b]{0.18\linewidth}
\includegraphics[width=1\linewidth]{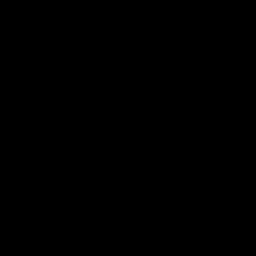} \\ \\
\includegraphics[width=1\linewidth]{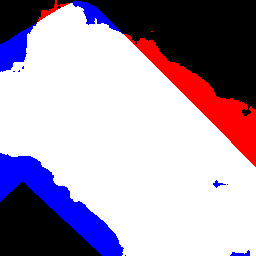}
\end{minipage}
}
\caption{Change detection results of DED and EDED on the NJDS in the first row and SYSU in the second row. Red areas denote false positives and blue areas denote false negatives. (a) T1 image. (b) T2 image. (c) Ground truth image. (d) DED result. (e) EDED result.}
\label{Fig:backbone_ablation}
\end{figure*}

EDED outperforms DED in ICCD and MVBCD since DED cannot segment all types of changed objects in the former scenario and confuses real changes and false positives of spatial differences in the latter scenario, as shown in Figure \ref{Fig:backbone_ablation}. The first row shows that EDED can correctly identify the spatial differences caused by different imaging angles as unchanged areas, while DED incorrectly identifies these spatial differences as changed areas, resulting in false positives. In the second row, EDED can detect most of the changed areas which contain multiple categories of objects accurately, while the result of DED has many false negatives.

\begin{table}[!ht]
\caption{Ablation Study of EDED Backbone on SYSU, LEVIR-CD, and NJDS Datasets. The Best Values Are Highlighted in Bold in Each Dataset.
}
\label{backbone_ablation}
\centering
\begin{tabular}{lccccc}
\toprule
Backbone & Dataset & P (\%) & R (\%) & F1 (\%) & IoU (\%) \\
\midrule
MESD & SYSU & 85.12 & 74.56 & 79.49 & 65.96\\
DED & SYSU & \textbf{85.66} & 75.84 & 80.45 & 67.30\\
EDED & SYSU & 85.46 & \textbf{78.24} & \textbf{81.69} & \textbf{69.05}\\
\midrule
MESD & LEVIR-CD & 93.01 & 89.44 & 91.19 & 83.80\\
DED & LEVIR-CD & 92.88 & 90.67 & 91.76 & 84.78 \\
EDED & LEVIR-CD & \textbf{93.09} & \textbf{91.32} & \textbf{92.20} & \textbf{85.52}\\
\midrule
MESD & NJDS & 78.04 & 63.57 & 70.07 & 53.92\\
DED & NJDS & 76.77 & 63.72 & 69.64 & 53.42\\
EDED & NJDS & \textbf{80.21} & \textbf{67.94} & \textbf{73.57} & \textbf{58.19}\\
\bottomrule
\end{tabular}
\end{table}

Table \ref{module_ablation} shows the ablation experiment results of HCU and TFAM on the LEVIR-CD dataset. Note that when comparing the model using conventional convolution and the model using HCU, the channel size of features in the latter model is twice that of the former model, making the number of parameters and computation of the two models consistent. The results show that HCU and TFAM are effective for binary change detection from different aspects. The first and second rows show that with consistent parameters and computation, using HCU can make the model more efficient and improve the performance; the first and third rows show that using TFAM can focus on the relatively important parts between bitemporal features and effectively fuse bitemporal features, thus improving the performance; the fourth row shows that using HCU and TFAM at the same time can further improve the performance. The above results indicate that HCU and TFAM can improve the performance, and using them together can further improve the change detection ability. This ablation study shows that HCU and TFAM are better choices for feature extraction and feature fusion on change detection task, respectively.

\begin{table}[!ht]
\caption{Ablation Study of Half Convolution Unit and Temporal Fusion Attention Module on LEVIR-CD. We Use EDED as The Backbone. Common Denotes the Basic Convolution Unit.}
\label{module_ablation}
\centering
\begin{tabular}{ccccccc} 
\toprule
Convolution Unit & Fusion & P (\%) & R (\%) & F1 (\%) & IoU (\%) \\
\midrule
Common & Add & 92.12 & 90.27 & 91.19 & 83.80 \\
HCU & Add & 93.09 & 91.32 & 92.20 & 85.52 \\
Common & TFAM & 92.87 & 91.26 & 92.16 & 85.46 \\
HCU & TFAM & 93.07 & 91.61 & 92.33 & 85.76 \\
\bottomrule
\end{tabular}
\end{table}

\subsection{Experimental Results}

\subsubsection{Intraclass Change Detection}
\label{section:4.4.3}

This subsection presents the results of comparing SGSLN with other change detection models on the ICCD task with two datasets (CDD and SYSU datasets).

The accuracy comparison results on the CDD dataset are presented in Table \ref{cdd_result}. It shows that SGSLN/512 surpasses all the compared models and achieves the highest IoU (0.9563) and F1-score (0.9777) on the CDD dataset. Compared with the second-best method (TransUNetCD), SGSLN/512 increases the F1-score by 0.6\%.

\begin{table}[!ht]
\caption{Accuracy Comparison on the CDD Dataset. The Best Values Are Highlighted in Bold.}
\label{cdd_result}
\centering
\begin{tabular}{lcccc}
\toprule
Methods & P (\%) & R (\%) & F1 (\%) & IoU (\%) \\
\midrule
FC-EF~\cite{fcef} & 60.90 & 58.30 & 59.57 & 42.42 \\
FC-Siam-Diff~\cite{fcef} & 76.20 & 57.30 & 65.41 & 48.60 \\
FC-Siam-Conc~\cite{fcef} & 70.90 & 60.30 & 65.17 & 48.34 \\
UNet++\_MSOF~\cite{etecd} & 86.68 & 76.53 & 81.29 & 68.48\\
IFN~\cite{dsifn} & 90.56 & 70.18 & 79.08 & 65.40 \\
BiT~\cite{bit} & 96.19 & 93.99 & 95.08 & 90.62 \\
SNUNet~\cite{snu} & 96.30 & 96.20 & 96.25 & 92.77\\ 
TransUNetCD~\cite{transunetcd} & 96.93 & \textbf{97.42} & 97.17 & 94.50 \\
SGSLN/128 & 94.79 & 92.76 & 93.76 & 88.26 \\
SGSLN/256 & 96.66 & 95.82 & 96.24 & 92.75 \\
SGSLN/512 & \textbf{98.25} & 97.29 & \textbf{97.77} & \textbf{95.63} \\
\bottomrule
\end{tabular}
\end{table}

The accuracy comparison results on the SYSU dataset are summarized in Table \ref{sysu_result}. Since the semantic information of the changed objects on the SYSU dataset is vague and contains objects of multiple categories, other change detection methods fail to accurately detect the changed objects. In contrast, SGSLN can detect the changed objects of all categories by using bitemporal semantic features to determine the changed areas of all changed objects. The accuracy comparison results show that SGSLN/512 achieves the highest IoU (0.7105) and F1-score (0.8307) on this dataset, surpassing all other models by a large margin. SGSLN/512, SGSLN/256 and SGSLN/128 all outperform other change detection models, increasing the F1-score by 2.04\%, 1.35\% and 0.07\% compared with the second-best model (DARNet), respectively.

\begin{table}[!ht]
\caption{Accuracy Comparison on the SYSU Dataset. The Best Values Are Highlighted in Bold.}
\label{sysu_result}
\centering
\begin{tabular}{lcccc}
\toprule
Methods & P (\%) & R (\%) & F1 (\%) & IoU (\%) \\
\midrule
FC-EF~\cite{fcef} & 74.32 & 75.84 & 75.07 & 60.09 \\
FC-Siam-Diff~\cite{fcef} & \textbf{89.13} & 61.21 & 72.58 & 56.96 \\
FC-Siam-Conc~\cite{fcef} & 82.54 & 71.03 & 76.35 & 61.75 \\
IFN~\cite{dsifn} & 79.59 & 75.58 & 77.53 & 63.31 \\
STANet~\cite{sta} & 70.76 & \textbf{85.33} & 77.36 & 63.09 \\
BiT~\cite{bit} & 81.14 & 76.48 & 78.74 & 64.94 \\
SNUNet~\cite{snu} & 78.26 & 76.30 & 77.27 & 62.96 \\
DARNet~\cite{darn} & 83.04 & 79.11 & 81.03 & 68.11 \\
SGSLN/128 & 82.48 & 79.77 & 81.10 & 68.21 \\
SGSLN/256 & 83.28 & 81.50 & 82.38 & 70.04 \\
SGSLN/512 & 84.76 & 81.45 & \textbf{83.07} & \textbf{71.05} \\
\bottomrule
\end{tabular}
\end{table}

\begin{figure}[!ht] 
\centering
\subfigure[]{
\begin{minipage}[b]{0.213\linewidth}
\includegraphics[width=1.1\linewidth]{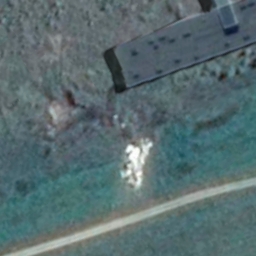} \\ \\
\includegraphics[width=1.1\linewidth]{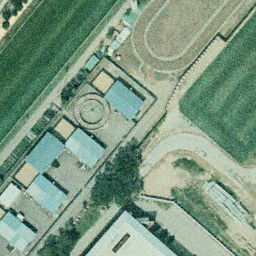}
\end{minipage}
}
\subfigure[]{
\begin{minipage}[b]{0.213\linewidth}
\includegraphics[width=1.1\linewidth]{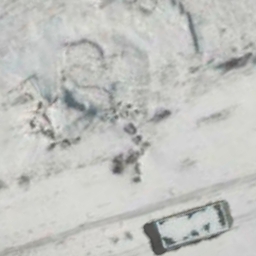} \\ \\
\includegraphics[width=1.1\linewidth]{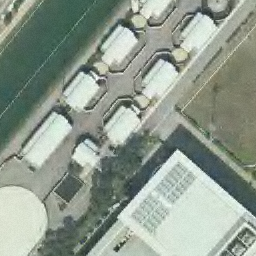}
\end{minipage}
}
\subfigure[]{
\begin{minipage}[b]{0.213\linewidth}
\includegraphics[width=1.1\linewidth]{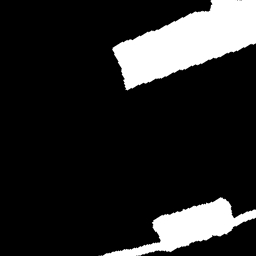} \\ \\
\includegraphics[width=1.1\linewidth]{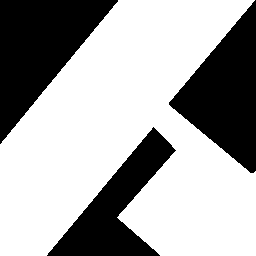}
\end{minipage}
}
\subfigure[]{
\begin{minipage}[b]{0.213\linewidth}
\includegraphics[width=1.1\linewidth]{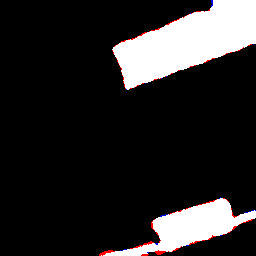} \\ \\
\includegraphics[width=1.1\linewidth]{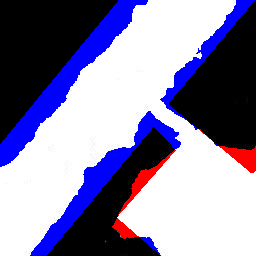}
\end{minipage}
}
\caption{Sample inference results of SGSLN/512 on the ICCD. The results on the CDD and SYSU datasets are shown in the first and second rows, respectively. Red areas denote false positives and blue areas denote false negatives. (a) T1 image. (b) T2 image. (c) Ground truth image. (d) SGSLN/512 result.}
\label{Fig:ic_result}
\end{figure}

The inference results of the test set of the CDD and SYSU datasets are shown in Figure \ref{Fig:ic_result}. The results show that SGSLN performs excellently on the two datasets. Under the strong interference of seasonal changes and illumination in CDD and the unclear semantic information of the changed objects in SYSU, SGSLN/512 still detects various types of changed objects with binary labels. 

\subsubsection{Single-View Building Change Detection}
\label{section:4.4.2}

This subsection presents the results of comparing SGSLN with other change detection models on the SVBCD task with three datasets (WHU, LEVIR-CD, and LEVIR-CD+ datasets).

The accuracy comparison results on the WHU dataset are shown in Table \ref{whu_result}. It shows that SGSLN/512 achieves the highest IoU (0.9022) and F1-score (0.9486) on this dataset, outperforming all other change detection models. At the same time, SGSLN/256 also outperforms all other models with a smaller size. Compared with the second-best method (FCCDN), SGSLN/512 and SGSLN/256 increase the F1-score by 1.12\% and 0.93\%, respectively.

\begin{table}[!ht]
\caption{Accuracy Comparison on the WHU Dataset. The Best Values Are Highlighted in Bold.}
\label{whu_result}
\centering
\begin{tabular}{lcccc}
\toprule
Methods & P (\%) & R (\%) & F1 (\%) & IoU (\%) \\
\midrule
FC-Siam-Diff~\cite{fcef} & 84.73 & 87.31 & 86.00 & 75.44 \\
FC-Siam-Conc~\cite{fcef} & 78.86 & 78.64 & 78.75 & 64.95\\ 
UNet++\_MSOF~\cite{etecd} & 91.96 & 89.40 & 90.66 & 82.92 \\
DTCDSCN~\cite{dtcdstn} & 63.92 & 82.30 & 71.95 & 56.19 \\
IFN~\cite{dsifn} & 91.44 & 89.75 & 90.59 & 82.79 \\
STANet~\cite{sta} & 79.37 & 85.50 & 82.32 & 69.95 \\
SNUNet~\cite{snu} & 85.60 & 81.49 & 83.49 & 71.67 \\
BiT~\cite{bit} & 86.64 & 81.48 & 83.98 & 72.39 \\
TransUNetCD~\cite{transunetcd} & 93.59 & 89.60 & 91.55 & 84.42 \\
FCCDN~\cite{fccdn} & \textbf{96.39} & 91.24 & 93.74 & 88.23 \\
SGSLN/128 & 93.52 & 89.91 & 91.68 & 84.64 \\
SGSLN/256 & 96.28 & 93.11 & 94.67 & 89.88 \\
SGSLN/512 & 96.11 & \textbf{93.64} & \textbf{94.86} & \textbf{90.22} \\
\bottomrule
\end{tabular}
\end{table}

The accuracy comparison results on the LEVIR-CD dataset are summarized in Table \ref{levircd_result}. It shows that SGSLN/512 outperforms all other models, achieving the highest IoU (0.8576) and F1-score (0.9233) on this dataset. Compared with the second-best method (FCCDN), SGSLN/512 increases the F1-score by 0.08\%.

\begin{table}[!ht]
\caption{Accuracy Comparison on the LEVIR-CD Dataset. The Best Values Are Highlighted in Bold.}
\label{levircd_result}
\centering
\begin{tabular}{lcccc}
\toprule
Methods & P (\%) & R (\%) & F1 (\%) & IoU (\%) \\
\midrule
FC-EF~\cite{fcef} & 86.91 & 80.17 & 83.40 & 71.53 \\
FC-Siam-Diff~\cite{fcef} & 89.53 & 83.31 & 86.31 & 75.91 \\
FC-Siam-Conc~\cite{fcef} & 91.99 & 76.77 & 83.69 & 71.96\\ 
DTCDSCN~\cite{dtcdstn} & 88.53 & 86.83 & 87.67 & 78.05\\ 
IFN~\cite{dsifn} & \textbf{94.02} & 82.93 & 88.13 & 78.77 \\
STANet~\cite{sta} & 83.81 & 91.00 & 87.26 & 77.39 \\
BiT~\cite{bit} & 89.24 & 89.37 & 89.30 & 80.68 \\
SNUNet~\cite{snu} & 89.18 & 87.17 & 88.16 & 78.83 \\
TransUNetCD~\cite{transunetcd} & 92.43 & 89.82 & 91.11 & 83.67\\
FCCDN~\cite{fccdn} & 92.96 & 91.55 & 92.25 & 85.61 \\
SGSLN/128 & 91.79 & 90.21 & 91.00 & 83.48 \\
SGSLN/256 & 92.71 & 91.17 & 91.93 & 85.07 \\
SGSLN/512 & 93.07 & \textbf{91.61} & \textbf{92.33} & \textbf{85.76} \\
\bottomrule
\end{tabular}
\end{table}

Table \ref{levircd+_result} presents the accuracy comparison results on the LEVIR-CD+ dataset. It shows that SGSLN/512 achieves the highest IoU (0.8414) and F1-score (0.9139) on this dataset, surpassing other models by a large margin. SGSLN/512, SGSLN/256 and SGSLN/128 all outperform other change detection models, increasing the F1-score by 5.12\%, 4.16\% and 3.66\% compared with the second-best model (Intelligent-BCD), respectively. Since LEVIR-CD+ adds more hard samples on the basis of LEIVR-CD, the performance of the same model on LEVIR-CD+ has a significant decline compared with that on LEVIR-CD, such as BiT and STANet with declines of 6.51\% and 7.99\% in the F1-score metric, while SGSLN/512, SGSLN/256 and SGSLN/128 only have declines of 0.94\%, 1.03\% and 1.05\% in the F1-score metric. This shows that SGSLN can resist the interference of building shadows and dense building distribution better than other change detection methods, and thus achieve superior performance in the more difficult SVBCD task.

\begin{table}[!ht]
\caption{Accuracy Comparison on the LEVIR-CD+ Dataset. The Best Values Are Highlighted in Bold.}
\label{levircd+_result}
\centering
\begin{tabular}{lcccc}
\toprule
Methods & P (\%) & R (\%) & F1 (\%) & IoU (\%) \\
\midrule
U-Net~\cite{unet} & 93.20 & 79.60 & 85.86 & 75.23 \\
AttU-Net~\cite{aunet} & 93.50 & 79.60 & 85.99 & 75.43 \\
FC-EF~\cite{fcef} & 61.30 & 72.61 & 66.48 & 49.79 \\
FC-Siam-Diff~\cite{fcef} & 74.97 & 72.04 & 73.48 & 58.07 \\
FC-Siam-Conc~\cite{fcef} & 66.24 & 81.22 & 72.97 & 57.44 \\
UNet++\_MSOF~\cite{etecd} & 85.90 & 67.10 & 75.34 & 60.44 \\
DTCDSCN~\cite{dtcdstn} & 80.36 & 75.03 & 77.60 & 63.40 \\
STANet~\cite{sta} & 74.62 & 84.54 & 79.27 & 65.66 \\
BiT~\cite{bit} & 82.74 & 82.85 & 82.79 & 70.64 \\
Intelligent-BCD~\cite{ibcd} & \textbf{93.80} & 79.90 & 86.29 & 75.89 \\
SGSLN/128 & 90.74 & 89.18 & 89.95 & 81.74 \\
SGSLN/256 & 91.30 & 90.50 & 90.90 & 83.32 \\
SGSLN/512 & 92.20 & \textbf{90.59} & \textbf{91.39} & \textbf{84.14} \\
\bottomrule
\end{tabular}
\end{table}

\begin{figure}[!ht] 
\centering
\subfigure[]{
\begin{minipage}[b]{0.213\linewidth}
\includegraphics[width=1.1\linewidth]{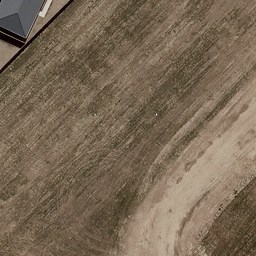} \\ \\
\includegraphics[width=1.1\linewidth]{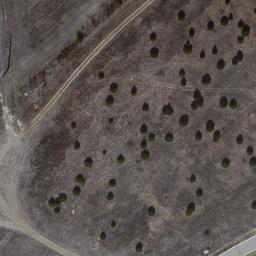} \\ \\
\includegraphics[width=1.1\linewidth]{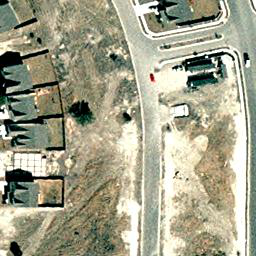}
\end{minipage}
}
\subfigure[]{
\begin{minipage}[b]{0.213\linewidth}
\includegraphics[width=1.1\linewidth]{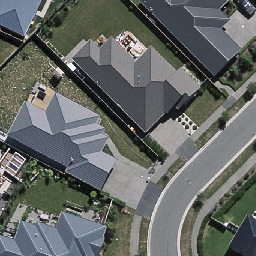} \\ \\
\includegraphics[width=1.1\linewidth]{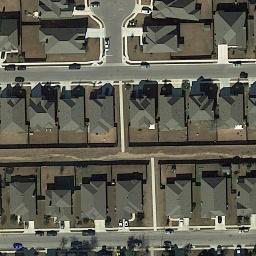} \\ \\
\includegraphics[width=1.1\linewidth]{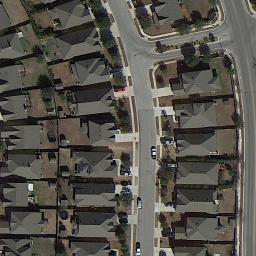}
\end{minipage}
}
\subfigure[]{
\begin{minipage}[b]{0.213\linewidth}
\includegraphics[width=1.1\linewidth]{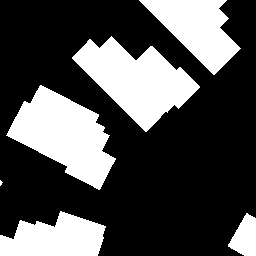} \\ \\
\includegraphics[width=1.1\linewidth]{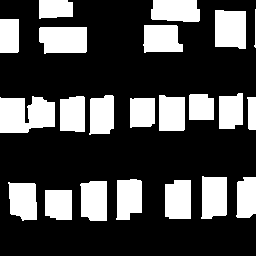} \\ \\
\includegraphics[width=1.1\linewidth]{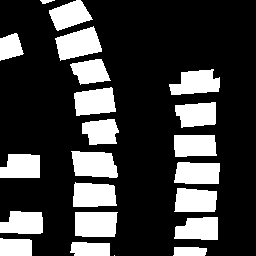}
\end{minipage}
}
\subfigure[]{
\begin{minipage}[b]{0.213\linewidth}
\includegraphics[width=1.1\linewidth]{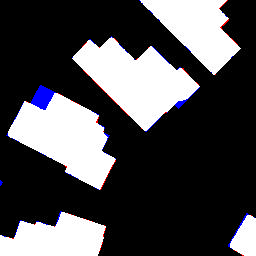} \\ \\
\includegraphics[width=1.1\linewidth]{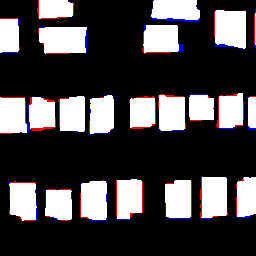} \\ \\
\includegraphics[width=1.1\linewidth]{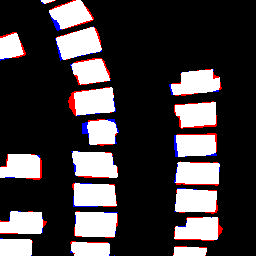}
\end{minipage}
}
\caption{Sample inference results of SGSLN/512 on the SVBCD. The results on WHU, LEVIR-CD, and LEVIR-CD+ datasets are shown in the first, second and third rows, respectively. Red areas denote false positives and blue areas denote false negatives. (a) T1 image. (b) T2 image. (c) Ground truth image. (d) SGSLN/512 result.}
\label{Fig:sv_result}
\end{figure}

We show some inference results of the test set of the WHU, LEVIR-CD, and LEVIR-CD+ datasets in Figure \ref{Fig:sv_result}. SGSLN/512 detects almost all the changed buildings in the three datasets. With the influence of building shadows and dense distribution of changed buildings, SGSLN/512 can still detect the regions and edges of changed buildings well. Note that there is a false change in the change mask of SGSLN/512 in the WHU dataset, as indicated by the blue part of the change mask in the first line. However, the building in the posttemporal image is under construction, which indicates building changes in this area. This is a frequent issue in this dataset as discussed in~\cite{lfc}.

\subsubsection{Multiview Building Change Detection}
\label{section:4.4.4}

This subsection presents the results of comparing SGSLN with other change detection models on the MVBCD task with the NJDS dataset.

Table \ref{njds_result} summarizes the accuracy comparison results on the NJDS dataset. It shows that SGSLN/512 achieves the highest IoU (0.5582) and F1-score (0.7165) on this dataset, surpassing all other models by a large margin. SGSLN/512 and SGSLN/256 both outperform other change detection models, increasing the F1-score by 4.82\% and 3.76\% compared with the second-best one (SFCCD), respectively.

\begin{figure}[!ht] 
\centering
\subfigure[]{
\begin{minipage}[b]{0.213\linewidth}
\includegraphics[width=1.1\linewidth]{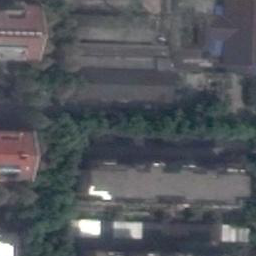} 
\end{minipage}
}
\subfigure[]{
\begin{minipage}[b]{0.213\linewidth}
\includegraphics[width=1.1\linewidth]{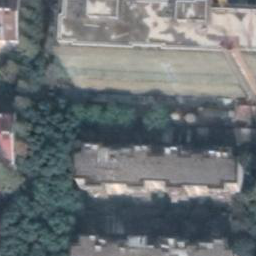}
\end{minipage}
}
\subfigure[]{
\begin{minipage}[b]{0.213\linewidth}
\includegraphics[width=1.1\linewidth]{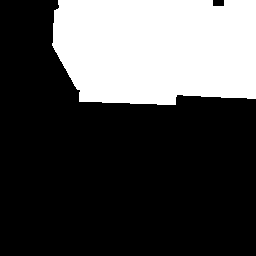}
\end{minipage}
}
\subfigure[]{
\begin{minipage}[b]{0.213\linewidth}
\includegraphics[width=1.1\linewidth]{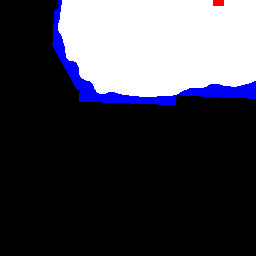}
\end{minipage}
}
\caption{Sample inference results of SGSLN/512 on the MVBCD. Red areas denote false positives and blue areas denote false negatives. (a) T1 image. (b) T2 image. (c) Ground truth image. (d) SGSLN/512 result.}
\label{Fig:njds_result}
\end{figure}

Figure \ref{Fig:njds_result} illustrates the inference results on the test set of the NJDS dataset. This shows that under the strong interference of spatial differences caused by the multiviews of both low-rise and high-rise buildings, SGSLN/512 can still accurately detect the change in high-rise and low-rise buildings and identify the spatial differences as unchanged.

\begin{table}[!ht]
\caption{Accuracy Comparison on the NJDS Dataset. The Best Values Are Highlighted in Bold.}
\label{njds_result}
\centering
\begin{tabular}{lcccc}
\toprule
Methods & P (\%) & R (\%) & F1 (\%) & IoU (\%) \\
\midrule
U-Net~\cite{unet} & 46.45 & 52.64 & 49.35 & 32.76 \\
AttU-Net~\cite{aunet} & 55.57 & 44.60 & 49.48 & 32.88 \\
PSPNet~\cite{psp} & 50.57 & 58.21 & 54.12 & 37.10 \\
DTCDSCN~\cite{dtcdstn} & 51.92 & 62.78 & 56.84 & 39.70 \\
IFN~\cite{dsifn} & 49.44 & 14.35 & 22.24 & 12.51 \\
MTU-Net~\cite{mtu} & 65.29 & 62.82 & 64.03 & 47.09 \\
SFCCD~\cite{njds} & 74.49 & 65.19 & 69.53 & 53.29 \\
SGSLN/128 & 71.01 & 59.40 & 64.69 & 47.81 \\
SGSLN/256 & 79.44 & 68.03 & 73.29 & 57.85 \\
SGSLN/512 & \textbf{79.92} & \textbf{69.51} & \textbf{74.35} & \textbf{59.18} \\
\bottomrule
\end{tabular}
\end{table}

\subsubsection{Efficiency Test}
\label{section:4.4.5}

\begin{table*}[!ht]
\caption{Efficiency Comparison on the WHU Dataset. The Best Values Are Marked with Bold Font. Params, FLOPs, IT, TB, TT, and F1 denote the number of parameters, computation costs, inference time with batch size = 1, training batch size with 12 GB memory, training time in 1 epoch, and F1-score, respectively.}
\label{efficiency}
\centering
\begin{tabular}{lcccccc}
\toprule
Name & Params (M) & FLOPs (G) & IT (s) & TB & TT (s) & F1 (\%) \\
\midrule
FC-EF~\cite{fcef} & 1.35 & 3.56 & \textbf{35} & 90 & 140 & 78.75\\
FC-Siam-Diff~\cite{fcef} & 1.54 & 5.30 & \textbf{35} & 65 & 140 & 86.00\\
FC-Siam-Conc~\cite{fcef} & 1.35 & 4.70 & \textbf{35} & 60 & 140 & 83.47\\
UNet++\_MSOF~\cite{etecd} & 9.05 & 34.01 & 40 & 14 & 170 & 90.66\\
IFN~\cite{dsifn} & 35.7 & 82.3 & 75 & 14 & 255 & 90.59\\
BiT~\cite{bit} & 3.55 & 67.8 & 40 & 45 & 145 & 83.98\\
FCCDN~\cite{fccdn} & 6.25 & 12.4 & 60 & 28 & 150 & 93.74\\
SGSLN/128 & \textbf{0.38} & \textbf{0.80} & 50 & \textbf{92} & \textbf{96} & 91.68\\
SGSLN/256 & 1.51 & 2.98 & 58 & 48 & 128 & 94.67\\
SGSLN/512 & 6.04 & 11.5 & 65 & 25 & 165 & \textbf{94.86}\\
\bottomrule
\end{tabular}
\end{table*}

This subsection reports the results of comparing SGSLN with other models in terms of parameters, computation, and accuracy. We conduct an efficiency test on the WHU dataset using the same implementation details as described in Section \ref{section:4.3.2}. Table \ref{efficiency} summarizes the results of the efficiency comparison. SGSLN/128 achieves an F1-score of 0.9168 with only 0.381 M parameters, 0.8045 FLOP computation, and 96 seconds of training time, surpassing other change detection models except FCCDN on F1-score. SGSLN/256 and SGSLN/512 surpass all the compared change detection models in F1-score with relatively low parameters, computation costs, inference time, and training time. This demonstrates that SGSLN can achieve superior performance with relatively low computation complexity and  high efficiency.

\section{Discussion}

\subsection{Exchanging Position}

The channel exchange module between the dual encoders lightens the EDED backbone. Without the channel exchange module, the model structure would resemble the DED architecture. It is not solely responsible for fusing bitemporal features as TFAM, its key contribution lies in the position of feature exchange. The channel exchange module makes the encoder features after the exchange have bitemporal semantic features of bitemporal images to determine changed areas, while the encoder features before the exchange retain the spatial features of bitemporal image for subsequent localization of changed objects. Therefore, the position of the channel exchange module is a key point for the EDED backbone. The position of exchanging bitemporal features should ensure that the encoder features before this position have rich spatial features, while the encoder features after this position have rich semantic features. We choose to exchange bitemporal features at the output position of encoding block 3, as in this position, the EDED backbone has better performance.

Table \ref{exchanging_comparsion} shows the results of SGSLN/512 with different exchange positions, where exchanging bitemporal features at the position of encoding block 3 can make SGSLN/512 perform best. We argue that exchanging bitemporal features at other positions makes the model perform worse due to the following reasons: (1) If exchanging bitemporal features at the position of encoding block 1 or 2, there are too few spatial features for subsequent localization of changed objects, leading to inaccurate detection of changed objects; and (2) If exchanging bitemporal features at the position of encoding block 4 or 5, the encoder features after the exchange have lost too much information due to downsampling many times, which makes it difficult for the model to detect all changed areas, and the encoder features are too small to detect small-scale changed areas. Therefore, exchanging features at the position of encoding block 3 can ensure that encoder features before the exchange have sufficient spatial features, while encoder features after the exchange have sufficient semantic features to detect all changed areas.

\begin{table}[ht!]
\caption{Accuracy of Different Exchanging Positions for SGSLN on WHU. The Best Values Are Highlighted in Bold.}
\label{exchanging_comparsion}
\centering
\begin{tabular}{ccccc}
\toprule
Position & P (\%) & R (\%) & F1 (\%) & IoU (\%) \\
\midrule
1 & 92.96 & 92.47 & 92.71 & 86.42 \\
2 & 95.06 & 92.12 & 93.57 & 87.91 \\
3 & \textbf{96.11} & \textbf{93.64} & \textbf{94.86} & \textbf{90.22} \\
4 & 94.78 & 92.89 & 93.83 & 88.37 \\
5 & 93.67 & 92.58 & 93.12 & 87.13 \\
\bottomrule
\end{tabular}
\end{table}

\subsection{Bitemporal Branches}

All triple branches in SGSLN generate change masks and be supervised by the same change label. What are the differences between the change masks produced by the bitemporal branches and the change mask generated by the fusion branch? How does the supervision of bitemporal branches affect the model's performance? We will discuss these two points in the following.

T1 branch utilize the semantic features of bitemporal images to identify changed area and spatial features of the T1 image to accurately localize T1 changed objects. Thus, the change mask generated by the T1 branch can detect T1 changed objects and coarse area of T2 change objects. Similarly, the T2 branch can detect T2 changed objects and coarse area of T1 change objects. The fusion branch then fuse bitemporal features and precisely detect bitemporal changed objects.

\begin{figure*}[!ht] 
\centering
\subfigure[]{
\begin{minipage}[b]{0.14\linewidth}
\includegraphics[width=1.1\linewidth]{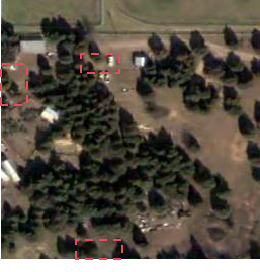} 
\end{minipage}
}
\subfigure[]{
\begin{minipage}[b]{0.14\linewidth}
\includegraphics[width=1.1\linewidth]{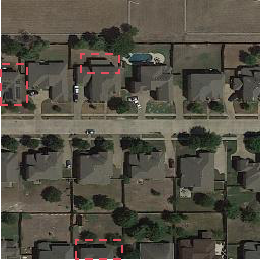}
\end{minipage}
}
\subfigure[]{
\begin{minipage}[b]{0.14\linewidth}
\includegraphics[width=1.1\linewidth]{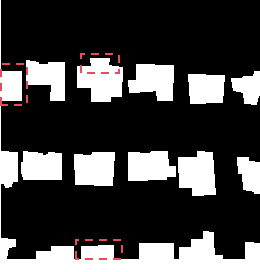}
\end{minipage}
}
\subfigure[]{
\begin{minipage}[b]{0.14\linewidth}
\includegraphics[width=1.1\linewidth]{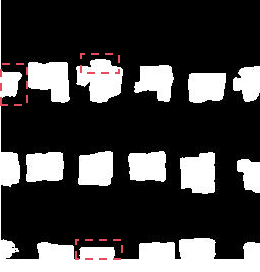}
\end{minipage}
}
\subfigure[]{
\begin{minipage}[b]{0.14\linewidth}
\includegraphics[width=1.1\linewidth]{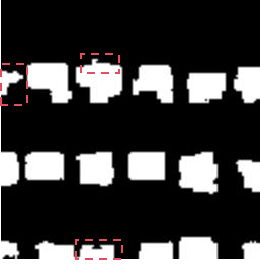}
\end{minipage}
}
\subfigure[]{
\begin{minipage}[b]{0.14\linewidth}
\includegraphics[width=1.1\linewidth]{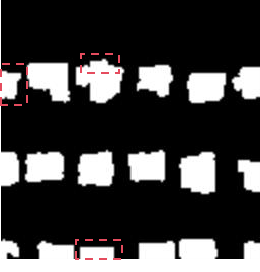}
\end{minipage}
}
\caption{Sample inference results of SGSLN/512 with results of triple branches on the LEVIR-CD+ dataset. (a) T1 image. (b) T2 image. (c) Ground truth image. (d) SGSLN/512 result in fusion branch. (e) SGSLN/512 result in T1 branch. (f) SGSLN/512 result in T2 branch.}
\label{Fig:levircd+_all_mask}
\end{figure*}

Figure \ref{Fig:levircd+_all_mask} illustrates the inference results of SGSLN/512 on the LEVIR-CD+ dataset. Numerous new buildings have been constructed in the T2 image, which means all the changed objects are distributed in the T2 image. The enclosed area within the red box demonstrates disparities among the change masks of triple branches. The change mask of T1 branch can only identify the rough area of changed buildings, while the T2 branch can precisely locate the changed buildings utilizing spatial features of the T2 image. The fusion branch result fuses features in bitemporal branches and precisely detect all changed buildings.

Supervising the bitemporal branches can enhance the training of the shallow layers. Since the gradient propagates from the final classification layer to the initial feature extraction layer in the deep network, the problem of gradient vanishing or exploding may occur in the propagation process, leading to the inefficiency of gradient propagation and poor features learned from lower layers~\cite{difficult_gradient_1,difficult_gradient_2,difficult_gradient_3,difficult_gradient_4,dsifn}. Supervision of bitemporal branches can reduce the distance of gradient back propagation and enable sufficient training of the shallow layers. Table \ref{supervision_ablation} presents the ablation of supervision branch on the LEVIR-CD+ dataset. Model with supervision of triple branches improves 0.8\% on F1-score and 1.34\% on IoU compared with model with supervision only on fusion branch, which demonstrates that supervision of bitemporal branches can enhance model training and improve performance of the model.

\begin{table}[!ht]
\caption{Ablation study of supervision branch of SGSLN/512 on LEVIR-CD+. Fusion branch means only fusion branch is supervised, and triple branches means fusion branch and bitemporal branches are all supervised.}
\label{supervision_ablation}
\centering
\begin{tabular}{lccccc} 
\toprule
Supervision Branch & P (\%) & R (\%) & F1 (\%) & IoU (\%) \\
\midrule
Fusion branch & 91.14 & 90.05 & 90.59 & 82.80 \\
Triple branches & 92.20 & 90.59 & 91.39 & 84.14 \\
\bottomrule
\end{tabular}
\end{table}

\subsection{Transferability}
We conducted a comparative analysis of transferability of SGSLN alongside other benchmark methods. All models are trained on the LEVIR-CD dataset, and then their accuracy was assessed on the WHU dataset. The resulting accuracy metrics are presented in Table \ref{transferability}.

The model trained on LEIVR-CD exhibits a significant decrease in performance when applied to WHU, primarily due to disparities in imaging conditions and image scenes between the LEIVR-CD and WHU datasets. SGSLN/512 outperforms other models in terms of F1-score and IoU on WHU, indicating that SGSLN/512 has the superior transferability compared to other methods. However, all pretrained models perform poorly on WHU. While SGSLN/512 achieves an F1-score of 0.9233 on LEIVR-CD, its F1-score on WHU drops to 0.5411. Training SGSLN/512 with WHU training data yields an F1-score of 0.9486 on the WHU test set, which means SGSLN/512 has considerable room for improvement in terms of transferability. In the future, we will focus on adjusting the model structure, enhancing training methods, and extending training data to bolster its transferability.

\begin{table}[!ht]
\caption{Accuracy Comparison on the Cross-domain Change Detection from LEVIR-CD to WHU.}
\label{transferability}
\centering
\begin{tabular}{lccccc} 
\toprule
Methods & P (\%) & R (\%) & F1 (\%) & IoU (\%) \\
\midrule
IFN~\cite{dsifn} & \textbf{78.92} & 21.73 & 34.08 & 20.54 \\
SNUNet~\cite{snu} & 69.84 & 38.29 & 49.46 & 32.86 \\
BiT~\cite{bit} & 64.38 & 33.83 & 44.35 & 28.50 \\
TransUNetCD~\cite{transunetcd} & 72.28 & 38.47 & 50.21 & 33.52 \\
FCCDN~\cite{fccdn} & 70.91 & 41.24 & 52.15 & 35.27 \\
SGSLN/512 & 73.46 & \textbf{42.83} & \textbf{54.11} & \textbf{37.09} \\
\bottomrule
\end{tabular}
\end{table}

\subsection{Expectations and Limitations}

Binary change detection is a fundamental task in change detection that aims to identify the changes of interest by comparing the remote sensing images of different time periods with binary labels. Various change detection methods have been proposed for this task, among which the MESD and DED architectures are widely adopted in network design. However, MESD suffers from the interference of bitemporal features in the fusion process, while DED is not suitable for ICCD and MVBCD scenarios because it fails to segment all types of changed objects in the former and confuses real changes with false positives of spatial differences in the latter.

Therefore, we propose an EDED backbone as a new strategy for binary change detection. EDED outperforms MESD and DED in the ICCD, SVBCD and MVBCD scenarios, as shown in Table \ref{backbone_ablation}. In ICCD, EDED can use bitemporal semantic features to determine changed areas, which are part of changed objects of all categories, and then use bitemporal spatial features to detect all types of changed objects. EDED increases the F1-score by 2.20\% and 1.24\% compared with MESD and DED on the SYSU dataset, respectively. In SVBCD, EDED can locate the changed buildings in each temporal using the bitemporal spatial features and accurately detect the region and edge parts of changed buildings. EDED increases the F1-score by 1.01\% and 0.44\% compared with MESD and DED on the LEVIR-CD dataset, respectively. In MVBCD, EDED can distinguish real changes from false changes caused by different imaging angles by using the bitemporal semantic features and accurately determine all the changed objects. EDED increases the F1-score by 3.50\% and 3.93\% compared with MESD and DED on the NJDS dataset, respectively. We expect the EDED backbone to be a new backbone for multiple binary change detection scenarios and achieve superior and robust performance in binary change detection.

Although EDED has superior performance compared with MESD and DED in binary change detection, it still has some drawbacks. EDED requires a large amount of multitemporal remote sensing images and binary labels for supervised training, in which the annotation of changed labels has high labor and time costs, leading to inadaptability of EDED in binary change detection with few or no labeled data. At the same time, EDED is oriented to binary change detection and cannot determine the category of changed objects in multitemporal remote sensing images, limiting its use in semantic change detection.

\section{Conclusion}

We propose an SGSLN model for binary change detection, which consists of an EDED backbone, HCUs and TFAMs. Specifically, we propose an EDED backbone as a new strategy for binary change detection, which solves the bitemporal feature interference problem in MESD by locating changed objects in each temporal separately and overcomes the limitations in ICCD and MVBCD in DED by using bitemporal semantic features to detect all types of changed objects in the former and distinguish spatial differences pseudochanges with real changes in the latter. We also propose a TFAM to fuse bitemporal features effectively by identifying the important parts between bitemporal features using temporal information and an HCU with 1/4 the number of parameters and computation of conventional convolution to achieve effective convolution and a lightweight model.

Experiments of SGSLN on the ICCD, SVBCD and MVBCD scenarios show that SGSLN achieves superior performance with high efficiency and outperforms all compared models. In ICCD scenarios, SGSLN can accurately determine various types of changed objects using bitemporal semantic features of bitemporal remote sensing images. In SVBCD scenarios, SGSLN can accurately locate changed buildings by spatial localization of changed objects in each temporal and temporal fusion of bitemporal features. In MVBCD scenarios, SGSLN can use bitemporal semantic features to distinguish spatial differences caused by multiviews of objects with real changes. We expect SGSLN to serve as a baseline for binary change detection, exploring its potential in more diverse change detection scenarios and achieving accurate and robust performance in change detection.

\section*{Acknowledgment}
The authors are grateful to the High-Performance Computing Center, Nanjing University, Nanjing, China, for their help with GPU resources. They would also like to thank the editor and the anonymous reviewers for their constructive comments.

\ifCLASSOPTIONcaptionsoff
  \newpage
\fi

\bibliographystyle{IEEEtranN}
\bibliography{paper.bib}

\end{document}